\documentclass[runningheads]{llncs}
\usepackage{graphicx}
\usepackage{amssymb}
\usepackage{amsmath}
\usepackage{caption}
\usepackage{subcaption}
\usepackage{blindtext}
\usepackage{etoolbox}
\usepackage{multirow}
\usepackage{color, colortbl}
\definecolor{Gray}{gray}{0.9}

\newcommand{\repthanks}[1]{\textsuperscript{\ref{#1}}}
\makeatletter
\patchcmd{\maketitle}
  {\def\thanks}
  {\let\repthanks\repthanksunskip\def\thanks}
  {}{}
\patchcmd{\@maketitle}
  {\def\thanks}
  {\let\repthanks\@gobble\def\thanks}
  {}{}
\newcommand\repthanksunskip[1]{\unskip{}}
\makeatother

\begin{document}
\title{Measuring Implicit Bias Using SHAP Feature Importance and Fuzzy Cognitive Maps}
\titlerunning{Measuring Implicit Bias Using SHAP and FCMs}
%
\author{Isel Grau\inst{1,2}\thanks{Equal contribution\protect\label{X}} \and Gonzalo N\'apoles\inst{3}\repthanks{X} \and Fabian Hoitsma \inst{3}, \\Lisa Koutsoviti Koumeri\inst{4} \and Koen Vanhoof \inst{4}}
\authorrunning{I. Grau et al.}
%
\institute{Information Systems Group, Eindhoven University of Technology, The Netherlands \and Eindhoven Artificial Intelligence Systems Institute, Eindhoven University of Technology, The Netherlands \and Department of Cognitive Science \& Artificial Intelligence, Tilburg University, The Netherlands \and Business Informatics Research Group, Hasselt University, Belgium.}

\maketitle              
\begin{abstract}
In this paper, we integrate the concepts of feature importance with implicit bias in the context of pattern classification. This is done by means of a three-step methodology that involves (i) building a classifier and tuning its hyperparameters, (ii) building a Fuzzy Cognitive Map model able to quantify implicit bias, and (iii) using the SHAP feature importance to active the neural concepts when performing simulations. The results using a real case study concerning fairness research support our two-fold hypothesis. On the one hand, it is illustrated the risks of using a feature importance method as an absolute tool to measure implicit bias. On the other hand, it is concluded that the amount of bias towards protected features might differ depending on whether the features are numerically or categorically encoded.
\end{abstract}
\section{Introduction}
\label{sec:introduction}
    
Fairness is a requirement that decision-makers are obliged to fulfill in any sector, as the law dictates that it is illegal to discriminate against so-called protected personal traits like gender or ethnicity. Therefore, decision-makers need to be able to ensure that their decision-making process is unbiased. Since Artificial Intelligence-based systems often assist decision-makers, such decision support systems are asked to be interpretable and transparent, which is a  challenging task for several reasons. First, fairness has multiple definitions depending on the case at hand, thus being difficult to quantify even if the related decision-making process is unbiased. Second, discrimination might be implicitly encoded in more than one feature of a dataset in non-linear ways and to different extents. Third, a trade-off between model accuracy and interpretability has been observed, thus making it difficult to understand how the decisions are made. This paper lies in the intersection of these challenges in an attempt to quantify implicit bias in pattern classification contexts.

A distinction is made between explicit and implicit bias. The former occurs when the decision-making outcome is influenced by protected features pre-defined by law~\cite{ntoutsi2020bias}, whereas the latter implies that the decision-making outcome is influenced by seemingly unbiased unprotected features that nevertheless reflect biased beliefs. An example is the redlining practice, where residents of minority neighborhoods receive less favorable treatment from financial institutions. This means that the place of residence can be used to implicitly discriminate against ethnic minorities~\cite{zhang2018fairness}. Another distinction is made between individual and group fairness. The first implies that similar individuals should be treated similarly and the second that different groups should be treated equally~\cite{mehrabi2021survey}. 

The majority of the existing bias measures reported in the literature focus on quantifying explicit bias against protected features. Group-based measures often involve probabilistic approaches that only consider the protected feature and the classification outcome effectively~\cite{mehrabi2021survey}, ignoring the rest of the information in the data. Individual-based approaches use distance metrics or regression-based tools that might not be sensitive enough to capture discrimination against a single sensitive feature, as shown in~\cite{Napoles2022c}. Therefore, we argue that explicit bias towards a single protected feature/group is a naive way to quantify fairness because discriminatory beliefs can find their way into the data through unprotected features that unexpectedly correlate with protected ones. 

Existing measures for implicit bias often rely on statistics using regression or correlation coefficients~\cite{vzliobaite2017measuring}. These approaches cannot easily capture non-linear and higher-order interactions. Other methods include classification rules~\cite{hajian2012methodology} and causal models~\cite{zhang2018fairness} which, despite being relatively interpretable,  mainly consider one-way interactions among a selected number of features. Moreover, they offer group-wise approaches~\cite{vzliobaite2017measuring} and require to assume which group is discriminated against, which might lead to reverse discrimination. In addition, existing approaches make poor use of interpretable machine learning, which is deemed an effective tool to enhance the transparency and fairness of models~\cite{fang2020achieving}. Feature importance methods are one way to illustrate how the model arrives at a certain prediction. However, in the context of fairness, these methods are mainly used to look for explicit bias since the features suspected of encoding bias are manually chosen by experts~\cite{fang2020achieving}. The works published in~\cite{meng2022interpretability,hickey2020fairness,alves2020fixout} rely on Shapley Additive Explanations (SHAP)~\cite{NIPS2017_7062}, which is one of the most explored model-agnostic explanation methods, for detecting and from there also mitigating explicit bias. However,~\cite{cesaro2019measuring} shows that it is difficult to get real insights on relationships among variables only by examining SHAP values.

In \cite{Napoles2022a}, the authors introduced a recurrent neural network-based model able to measure implicit bias with regard to seemingly neutral unprotected features. This model leverages Fuzzy Cognitive Maps (FCMs)\cite{Kosko1986}, a soft computing technique able to model the behavior of complex systems and perform what-if simulations. FCMs are able to capture higher-order associations, feedback loops and dependencies within the data. Therefore, we argue that they are a pertinent tool to capture the way that bias implicitly spreads and diffuses within data. This approach transforms the dataset into a fully connected graph, where each node corresponds to a problem feature. The nodes, or concepts, are connected using weighted edges denoting absolute pairwise correlations between features. Each neural concept is assigned an activation value representing its initial influence within the system. FCMs allow concepts to interact with each other by updating these activation values iteratively using a reasoning function. The final activation values after convergence are used as a proxy for implicit bias since the model considers all possible pathways through which bias can propagate through the system. The theoretical contribution of the work published in \cite{Napoles2022a} is a novel reasoning rule coupled with a normalization-like transfer function. These guarantee convergence to a unique fixed point regardless of the initial conditions or diverse point attractors by adjusting the influence of the reasoning rule's linear component. The main limitation of this model is that it relies on domain knowledge to activate the network.

In this paper, we design an experiment to illustrate that feature importance might hide implicit bias against protected features in decision-making problems. In consequence, we argue that feature importance should not be used as a direct proxy to discard bias in a dataset. This experiment is implemented in a three-step methodology that includes fitting a classifier and optimizing its hyperparameters, using our FCM model to quantify implicit bias, and running what-if simulations using SHAP feature importance to feed the FCM model. This initialization overcomes the limitation of setting the initial vector of the FCM model based on domain knowledge. Consequently, it offers a way to measure how an implicitly biased unprotected feature influences the prediction for an instance. An additional theoretical contribution of this paper is using clustering to automatically discover the groups describing a numeric feature, which allows computing the association between numeric and categorical features using Cram\'er's V~ coefficient \cite{cramer2016mathematical}. This strategy removes the need to define the groups manually while dealing with the limitations of determining the association between numeric and nominal features. Towards the end, we discuss the effect of discretizing features on the implicit bias analysis.

The structure of this paper is as follows. Section~\ref{sec:preliminaries} introduces the preliminaries concerning fuzzy cognitive mapping and the SHAP method used for computing feature importance. Section~\ref{sec:methodlogy} elaborates on the proposed methodology and the two-fold research hypothesis. Section~\ref{sec:simulations} presents the simulation results while Section~\ref{sec:conclusions} presents the concluding remarks.

\section{Preliminaries}
\label{sec:preliminaries}

This section will describe the building blocks of our methodology, namely the FCM model and the SHAP method for feature importance.

\subsection{Fuzzy Cognitive Maps}
\label{sec:preliminaries:fcm}

The classic FCM model introduced in \cite{Kosko1986} consists of a collection of meaningful neural entities called concepts that describe the modeled complex system. The interaction between these neurons is governed by a squared weight matrix such that $w_{ij} \in [-1,1]$. FCMs are knowledge-based recurrent neural networks, and as such, they perform an iterative reasoning process devoted to updating neurons' activation values given an initial condition. 

The traditional FCM model uses monotonically increasing transfer functions (such as the sigmoid and hyperbolic tangent functions) to ensure that neurons' activation values are in the desired bounded interval \cite{napoles2020}. Moreover, the recurrent process is closed, meaning that the state of the network in the current iteration solely depends on the previous state. These features cause a wide variety of problems ranging from saturation situations (where the activation values move towards the boundaries of the activation interval) to serious convergence problems. N\'apoles et al. \cite{Napoles2022a} expanded the quasi-nonlinear FCM model presented in \cite{Napoles2022b} and introduced a re-scaled transfer function to address these drawbacks. This model will be briefly described next.

Let $\mathbf{A}^{(t)}=( a_{1}^{(t)}, \ldots, a_{i}^{(t)}, \ldots, a_{m}^{(t)} )$ be the activation vector produced by an FCM such that $a_{i}^{(t)}$ is the activation value of the $i$-th neuron in the $t$-th iteration and $m$ is the number of neurons. Moreover, $\bar{\mathbf{A}}^{(t)}=(\bar{a}_{1}^{(t)}, $ $\ldots, \bar{a}_{i}^{(t)}, $ $\ldots, \bar{a}_{m}^{(t)})$ is the raw activation vector where $\bar{a}_{i}^{(t)}$ is the raw activation value of the $i$-th neuron in the current iteration. More explicitly, the vector $\bar{\mathbf{A}}^{(t)}$ is given by $\bar{\mathbf{A}}^{(t)} = \mathbf{A}^{(t)}\mathbf{W}$ where $\mathbf{W}_{m \times m}$ is the weight matrix. Equation \eqref{eq:reasoning} shows the quasi-nonlinear reasoning rule using a re-scaled transfer function,

\begin{equation}
\label{eq:reasoning}
\mathbf{A}^{(t+1)} = \phi f \left(\mathbf{A}^{(t)}\mathbf{W} \right) + (1 - \phi)\mathbf{A}^{(0)}
\end{equation}

\noindent such that $f(.): \mathbb{R}^m \rightarrow \mathbb{R}^m $ is defined as follows:

\begin{equation}
\label{eq:transfer-function}
f(\mathbf{X}) =
\begin{cases}
\frac{\mathbf{X}}{\lvert\lvert \mathbf{X} \rvert\rvert_2} & \text{if $X \neq \overrightarrow{0}$}\\
0 & \text{otherwise.}
\end{cases}
\end{equation}

\noindent where $||\cdot||_2$ represents for the Euclidean norm, and $0 \leq \phi \leq 1$ controls the nonlinearity of the reasoning rule. When $\phi=1$, the model performs as a closed system where the activation value of a neuron depends on the activation values of connected neurons in the previous iteration. When $0<\phi<1$, the model adds a linear component to the reasoning rule concerning the initial activation values of neurons \cite{Napoles2022a}. When $\phi=0$, the model narrows down to a linear regression where the initial activation values of neurons act as regressors \cite{Napoles2022b}.

The recurrent reasoning mechanism stops when either (i) the model converges to a fixed point or (ii) a maximal number of iterations $T$ is reached. In the quasi-nonlinear FCM model, we have the following states:

\begin{itemize}
  \item \textbf{Fixed point} $(\exists t_{\alpha} \in \{1,\dots,(T-1)\} : \mathbf{A}^{(t+1)}=\mathbf{A}^{(t)}, \forall i, \forall t \geq t_{\alpha})$: the FCM produces the same state vector after $t_{\alpha}$, thus $\mathbf{A}^{(t_{\alpha})}=\mathbf{A}^{(t_{\alpha}+1)}=\mathbf{A}^{(t_{\alpha}+2)}=\dots=\mathbf{A}^{(T)}$. If the fixed point is unique, the FCM model will produce the same state vector regardless of the initial conditions. \\
  \item \textbf{Limit cycle} $(\exists t_{\alpha},p,j \in \{1,\dots,(T-1)\}: \mathbf{A}^{(t+p)} = \mathbf{A}^{(t)},$ $ \forall i, \forall t \geq t_{\alpha})$: the FCM produces the same state vector periodically after the period $p$, thus $\mathbf{A}^{(t_{\alpha})}=a_i^{(t_{\alpha}+p)}=\mathbf{A}^{(t_{\alpha}+2p)}=\dots=\mathbf{A}^{(t_{\alpha}+jp)}$, where $t_{\alpha}+jp \leq T$. \\
  \item \textbf{Chaos}: the FCM produces different state vectors.
\end{itemize}

If $\phi=1$, the re-scaled FCM model is expected to converge to the unique fixed-point attractor provided that the weight matrix $W$ has an eigenvalue that is strictly greater in magnitude than the other eigenvalues and that the initial activation vector $\mathbf{A}^{(0)}$ has a nonzero component in the direction of an eigenvector associated with the dominant eigenvalue \cite{Napoles2022a}. If these conditions are fulfilled, the network will converge to the dominant eigenvalue. 

If $0 \leq \phi < 1$, the fixed point will depend on the initial conditions, thus allowing for what-if simulations. However, limit cycles can appear if $\mathbf{A}^{(t)}\mathbf{W} = \overrightarrow{0}$ for some $t$. Similarly, a cycle will be reached if the function $f(.)$ is evaluated on the discontinuity point $\overrightarrow{0}$ during the inference process. Overall, a limit cycle can appear if the following expression is fulfilled:

$$\left( \frac{\mathbf{A}^{(t+p-1)}}{\lvert\lvert \mathbf{A}^{(t+p-1)}\mathbf{W} \rvert\rvert_2} - \frac{\mathbf{A}^{(t-1)}}{\lvert\lvert \mathbf{A}^{(t-1)}\mathbf{W} \rvert\rvert_2} \right)\mathbf{W} = 0.$$

The previous equality holds when the vector resulting from the difference of the fractions is perpendicular to every column of $\mathbf{W}$ or such a difference is equal to the null vector. These situations might happen when

$$\frac{\mathbf{A}^{(t+v-1)}}{\lvert\lvert \mathbf{A}^{(t+v-1)}\mathbf{W} \rvert\rvert_2} = \frac{\mathbf{A}^{(t-1)}}{\lvert\lvert \mathbf{A}^{(t-1)}\mathbf{W} \rvert\rvert_2},$$

\noindent and

$$\mathbf{A}^{(t+p-1)} = \frac{\lvert\lvert \mathbf{A}^{(t+p-1)}\mathbf{W} \rvert\rvert_2}{\lvert\lvert \mathbf{A}^{(t-1)}\mathbf{W} \rvert\rvert_2} \mathbf{A}^{(t-1)}.$$

The above expression suggests that $\mathbf{A}^{(t+p-1)}$ is equal to $\mathbf{A}^{(t-1)}$ multiplied by a scalar factor, which is the quotient of the Euclidean norms. This means that if an activation vector is multiple of another activation for the same initial stimulus, then the cyclic behavior appears.

\subsection{Feature Importance and SHAP method}
\label{sec:preliminaries:shap}

Shapley Additive Explanations (SHAP) \cite{NIPS2017_7062} is a model-agnostic post-hoc method that computes feature importance as an approximation of Shapley values \cite{shapley1953value}. Shapley values come from the field of coalitional game theory and represent how much a feature brings in for a prediction, in addition to a given subset of features. Formally, a Shapley value $S_c$ represents the importance of the feature $c$ when included in the model $g$, that is to say:

\begin{equation}
    S_c = \sum_{B \subseteq C \setminus \{c\} } \frac{|B|!(|C|-|B|-1)!}{|C|!} \left (g(x_{B\cup \{c\}})-g(x_B) \right) 
\end{equation}

\noindent where $C$ represents the original feature set, $B$ stand for the possible subsets of $C \setminus \{c\}$, while $g(x_{B\cup\{c\}})$ denotes the prediction for the instance $x$ from the model $g$ when including all features in $B$ plus the feature $c$, and marginalizing over the rest of the features. Shapley values comply with the efficiency property, i.e., the sum of all feature contributions equals the difference between the prediction for $x$ and the average prediction. In other words, this means that the feature attribution can be aggregated for groups of features.

Shapley values can be calculated for a single instance or globally as an aggregation over all instances. However, this aggregation is computed over all possible combinations of feature subsets (or coalitions), therefore it can be computationally expensive. The SHAP implementation \cite{NIPS2017_7062} builds upon Shapley values theory and reframes it as an additive feature attribution method, i.e., a linear model. For example, the model-agnostic Kernel-SHAP method uses a local linear regression for estimating the values. Other model-specific implementations of SHAP include Tree-SHAP \cite{lundberg2020local2global}, which is optimized for decision trees, random forests, and gradient-boosted trees, by using the number of training examples traversing the tree to represent the background distributions.

\section{Methodology}
\label{sec:methodlogy}

\begin{figure}[!htbp]
    \centering
    \resizebox{\textwidth}{!}{
    \includegraphics{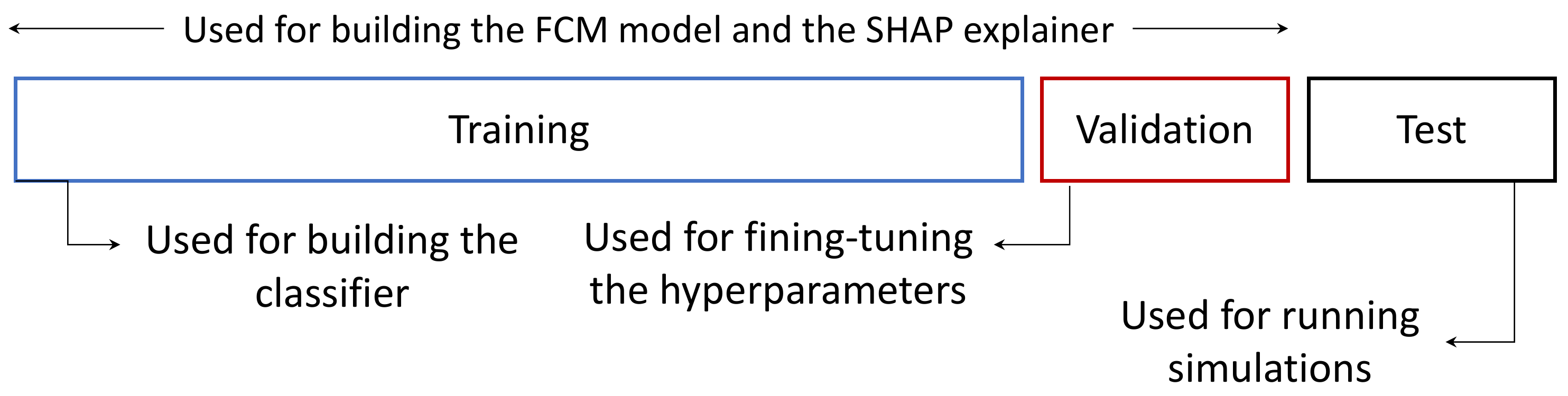}}
    \caption{Blueprint of the data usage in the proposed experiment.}
    \label{fig:methodology}
\end{figure}

This section presents our research methodology, which consists of the following steps (i) building a classifier, tuning its hyperparameters, (ii) building an FCM model able to quantify implicit bias, fitting the SHAP explainer, and (iii) running simulations. To build the classifier, we need a training dataset (70\%) and a separate validation dataset (20\%) to perform hyperparameter tuning. Both pieces of data can be combined to build the FCM model and fit the SHAP explainer once the classifier has been built. Finally, we can select some instances from the test set (10\%) for running simulations using the SHAP feature importance scores to activate the neurons in the FCM model. In that way, we can quantify how these feature importance scores translate to implicit bias against protected features. Figure \ref{fig:methodology} summarizes how the dataset is split toward obtaining the (stratified) training, validation, and test sets mentioned above.

The main hypothesis of our research is two-fold. On the one hand, we state that feature importance is not a suitable measure to determine bias against protected features. The fact that a protected feature is not regarded as relevant according to the SHAP values does not allow automatically concluding that the decisions are not biased towards that feature. On the other hand, we conjecture that the conclusions about the amount of bias against protected features might differ depending on how the feature is encoded.

\subsection{Building the Classification Model}
\label{sec:methodlogy:data}

In this paper, we will use a Random Forest (RF) classifier \cite{Breiman2001} as the core decision model. The aim is to investigate the extent to which protected features are regarded as important (as determined by the SHAP values). In order to ensure the reliability of the analysis, we need to ensure that RF produces the highest prediction rates possible. Therefore, it seems convenient to perform hyperparameter tuning using the training and validation sets.

The hyperparameters to be optimized are the number of estimators, the function used to measure the quality of a split, and the number of features to consider when looking for the best split. In our experiments, we will consider 100, 500, and 1000 estimators, while the maximum number of features can be determined as $\sqrt{m}$ or $log_2(m)$, with $m$ being the number of features. As for the split quality function, we will consider the Gini impurity and the Shannon information gain. Once the best parameter combination is determined, we will retrain the classifier using all data but the test set.

\subsection{Building the FCM-based Model}
\label{sec:methodlogy:fcm}

The next step in our methodology consists of creating an FCM model to quantify implicit bias based on the model proposed by N\'apoles et al. \cite{Napoles2022a}. 

In the knowledge-based network, each neuron represents a problem feature regardless of whether the feature is nominal or continuous since neurons will be activated using the SHAP values. The weights connecting the neurons denote the degree of association between the features in the dataset. In this step, we need to select the proper method to compute the association between two features based on whether these features are numerical or nominal. Let $F_i$ and $F_j$ denote two problem features denoted by neurons $C_i$ and $C_j$, respectively. The weight $w_{ij}$ connecting $C_i$ and $C_j$ will be determined as follows:

\begin{itemize}
    \item \textbf{Case 1}. Both $F_i$ and $F_j$ are numeric. In this case, the weight $w_{ij}$ is computed as the absolute Pearson's correlation \cite{rovine199714th}.\\
    \item \textbf{Case 2}. Both $F_i$ and $F_j$ are nominal. In this case, the weight $w_{ij}$ is computed using the Cram\'er's V~ coefficient \cite{cramer2016mathematical}.\\
    \item \textbf{Case 3}. Either $F_i$ or $F_j$ is numeric. In this case, we first transform the numerical feature into a nominal one by using the fuzzy $c$-means algorithm \cite{Bezdek1984}. The optimal number of clusters is determined using the fuzzy partition coefficient \cite{bezdek2013pattern}, which measures the amount of overlap among the fuzzy clusters. Once the numeric feature has been discretized, computing the weight $w_{ij}$ narrows down to the second case explained above.
\end{itemize}

More details on the automatic detection of categories describing a numerical feature are provided next. Firstly, the feature values $x_i$ are represented as symmetric tuples $(x_i,x_i)$ that can be represented in a plane. This suggests that the fuzzy $c$-means algorithm will discover $c$ fuzzy sets along the identity line, where each fuzzy cluster denotes a category.

The fuzzy component of this algorithm a membership is given by a matrix $\mathbf{U}_{n \times c}$ such that $n$ is the number of data points to be processed (i.e., the number of instances in the dataset). As such, $\mu_{ij} \in \mathbf{U}$ represents the degree to which the $i$-th data point belongs to the $j$-th fuzzy cluster. The algorithm returns a matrix of prototypes $\mathbf{Z}_{1 \times c}$ denoting the cluster centers. The fuzziness of fuzzy $c$-means is controlled by a fuzzification coefficient $\alpha \in[1,\infty]$ where larger values indicate more fuzziness. Equations \eqref{eqn:fcmeans_memfun} and \eqref{eqn:fcmeans_cj} display how to compute the membership values and the fuzzy prototypes, respectively:

\begin{equation}
\mu_{ij} = \frac{1}{\sum\limits_{l=1}^c \Big( \frac{\Vert x_i - z_j \Vert}{\Vert x_i - z_l \Vert} \Big)^{2/(\alpha-1)} }
\label{eqn:fcmeans_memfun}
\end{equation}
\begin{equation}
z_j = \frac{\sum\limits_{i=1}^n \mu_{ij}^\alpha \cdot x_i}{\sum\limits_{i=1}^n \mu_{ij}^\alpha}.
\label{eqn:fcmeans_cj}
\end{equation}

Since the number of categories needs to be discovered, we need to execute the clustering algorithms several times for different numbers of clusters (normally, from 2 to 10). The setting with the largest fuzzy partition coefficient is adopted to describe the numerical feature. This coefficient measures the amount of overlap between the fuzzy clusters and can be computed as follows:

\begin{equation}
\label{eq:fpc}
FPC=\frac1n\sum_{j=1}^c\sum_{i=1}^n{\left(\mu_\textit{ij}\right)}^{\alpha}.
\end{equation}

It is worth mentioning that the approach to handling numeric-nominal pairs of features is a contribution of this paper. In the method proposed in \cite{Napoles2022a}, the authors used the R-squared coefficient of determination~\cite{nagelkerke1991note} to quantify the percentage of variation in the numeric feature that the nominal one explains. While statistically sound, this strategy poses two issues. Firstly, it was assumed that such association is symmetric. Secondly, the results of analyzing the amount of bias against a feature might change depending on the encoding of that feature. However, automatically discovering groups associated with protected features (as done with the fuzzy $c$-means algorithm) could remove the subjectivity of defining the protected groups. Ultimately, the fuzzy clustering approach will help study the second hypothesis of our study.

\subsection{Initializing the FCM-based Model}
\label{sec:methodlogy:shap}

The final step of our methodology for fairness analysis notably differs from the model in \cite{Napoles2022a}, where selected neurons were randomly activated to perform the reasoning process depicted in Equation \eqref{eq:reasoning}. In our study, the initial activation vector $\mathbf{A}^{(0)}=( a_{1}^{(0)}, \ldots, a_{i}^{(0)}, \ldots, a_{m}^{(0)} )$ used the trigger reasoning will be unitized with the SHAP values for selected instances. Therefore, starting from feature importance, we will quantify how the associations among the variables increase the activation values of protected features. This can be done by exploring the final activation vector $\mathbf{A}^{(T)}=( a_{1}^{(0)}, \ldots, a_{i}^{(T)}, \ldots, a_{m}^{(T)} )$. Such an increase can be understood as implicit bias: the protected feature is not considered relevant by itself when making the decision; however, its patterns are encoded into unprotected features through correlations and associations.

\section{Simulations}
\label{sec:simulations}

In the numerical simulations, we use two datasets to illustrate the extent to which protected features are implicitly biasing the final decision of randomly selected individuals. First, we compute the global and local SHAP values after training a classifier. Second, we build an FCM model where weights are a square association matrix characterizing the interaction between the variables. In this model, the initial activation vector uses the local SHAP values of a randomly selected individual. Third, we quantify implicit bias related to specific individuals using w.r.t. interesting protected features.

\subsection{German Credit}

The first case study in our simulations concerns the German Credit dataset~\cite{Dua2019}. This binary classification dataset consists of 1000 credit applications, from which 700 are classified as good credit risk, while 300 are labeled as bad credit risk. Applicants are described by 20 qualitative and quantitative features. In this dataset, \textit{age} (F13), \textit{foreign worker} (F20) and \textit{gender} (F9) are considered to be protected and will be the center of our analysis.

Table \ref{table:association} shows the association values between all features and the protected ones. Notice that some unprotected features have rather strong associations with the protected ones. For example, the unprotected features employment since (F7), residence since (F11) and housing (F15) are strongly associated with \textit{age} (F13) which might be an indication that they implicitly encode \textit{age} bias. Table \ref{table:association} also displays the global SHAP values using a Random Forest as a classifier. This measure indicates that checking account (F1) and duration (F2) are key features when making the decision. It is worth mentioning that the weight matrix of our FCM model is symmetric, thus the matrix is diagonalizable and its eigenvalues are real with a dominant eigenvalue.

\begin{table}[!htb]
\centering
\caption{Association values between protected and unprotected features in the German Credit dataset. Global SHAP values provide information about feature importance using a Random Forest as a classifier.\\}
\label{table:association}
\begin{tabular}{l|l|c|c|c|c} 
\hline
\multicolumn{1}{c|}{\multirow{2}{*}{ID}} & \multicolumn{1}{c|}{\multirow{2}{*}{Features}} & \multicolumn{3}{c|}{Associates with} & \multirow{2}{*}{SHAP}  \\ 
\cline{3-5}
\multicolumn{1}{c|}{}                    & \multicolumn{1}{c|}{}                          & Gender & Age  & Foreign worker        &                       \\ 
\hline
F1                                       & Checking account                               & 0.03   & 0.08 & 0.08                  & 0.093                   \\
F2                                       & Duration                                       & 0.11   & 0.04 & 0.17                  & 0.035                   \\
F3                                       & Credit history                                 & 0.12   & 0.13 & 0.07                  & 0.023                   \\
F4                                       & Purpose                                        & 0.15   & 0.14 & 0.17                  & 0.019                   \\
F5                                       & Credit amount                                  & 0.08   & 0.03 & 0.04                  & 0.026                   \\
F6                                       & Savings account                                & 0.07   & 0.10 & 0.04                  & 0.026                   \\
F7                                       & Employment since                               & 0.22   & 0.37 & 0.08                  & 0.012                   \\
F8                                       & Installment rate                               & 0.13   & 0.06 & 0.13                  & 0.009                   \\
\cellcolor{Gray}F9                       & \cellcolor{Gray}Gender                        & 1.00   & 0.11 & 0.05                  & 0.015                   \\
F10                                      & Other debtors                                  & 0.01   & 0.02 & 0.12                  & 0.005                   \\
F11                                      & Residence since                                & 0.11   & 0.27 & 0.08                  & 0.009                   \\
F12                                      & Property                                       & 0.09   & 0.17 & 0.14                  & 0.017                   \\
\cellcolor{Gray}F13                      & \cellcolor{Gray}Age                            & 0.11   & 1.00 & 0.02                  & 0.013                   \\
F14                                      & Other installment                              & 0.05   & 0.03 & 0.04                  & 0.018                   \\
F15                                      & Housing                                        & 0.23   & 0.21 & 0.07                  & 0.017                   \\
F16                                      & Existing credits                               & 0.10   & 0.15 & 0.02                  & 0.004                   \\
F17                                      & Job                                            & 0.09   & 0.12 & 0.11                  & 0.010                   \\
F18                                      & People liable                                  & 0.20   & 0.13 & 0.07                  & 0.003                   \\
F19                                      & Telephone                                      & 0.07   & 0.09 & 0.10                  & 0.010                  \\
\cellcolor{Gray}F20                      & \cellcolor{Gray}Foreign worker                 & 0.05   & 0.02 & 1.00                  & 0.003                   \\
\hline
\end{tabular}

\end{table}

To study our two-fold hypothesis, we will select two randomly selected instances from the test set (one belonging to the good credit risk class and another belonging to the bad credit risk class). After verifying that instances have been correctly classified, we determine their associated SHAP values, which will be used to activate the FCM model. In that way, we can study how the implicit bias behaves given these feature importance scores. 

\begin{figure*}[!htb]
\center
    \begin{subfigure}{0.9\textwidth}
	\center
	\includegraphics[width=\textwidth, trim=0 2.2cm 0 0,clip]{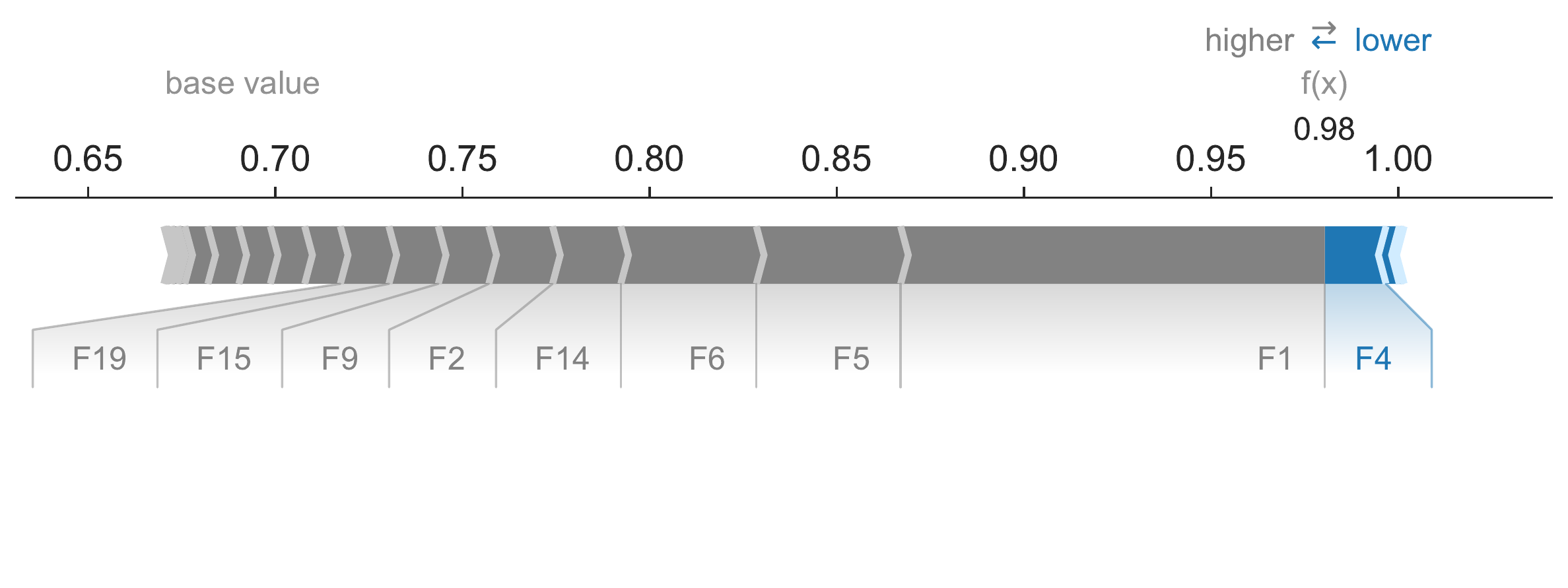}
	\caption{positive instance}
	\end{subfigure}
	\begin{subfigure}{0.9\textwidth}
	\center
	\includegraphics[width=\textwidth, trim=0 2.2cm 0 -0.5cm,clip]{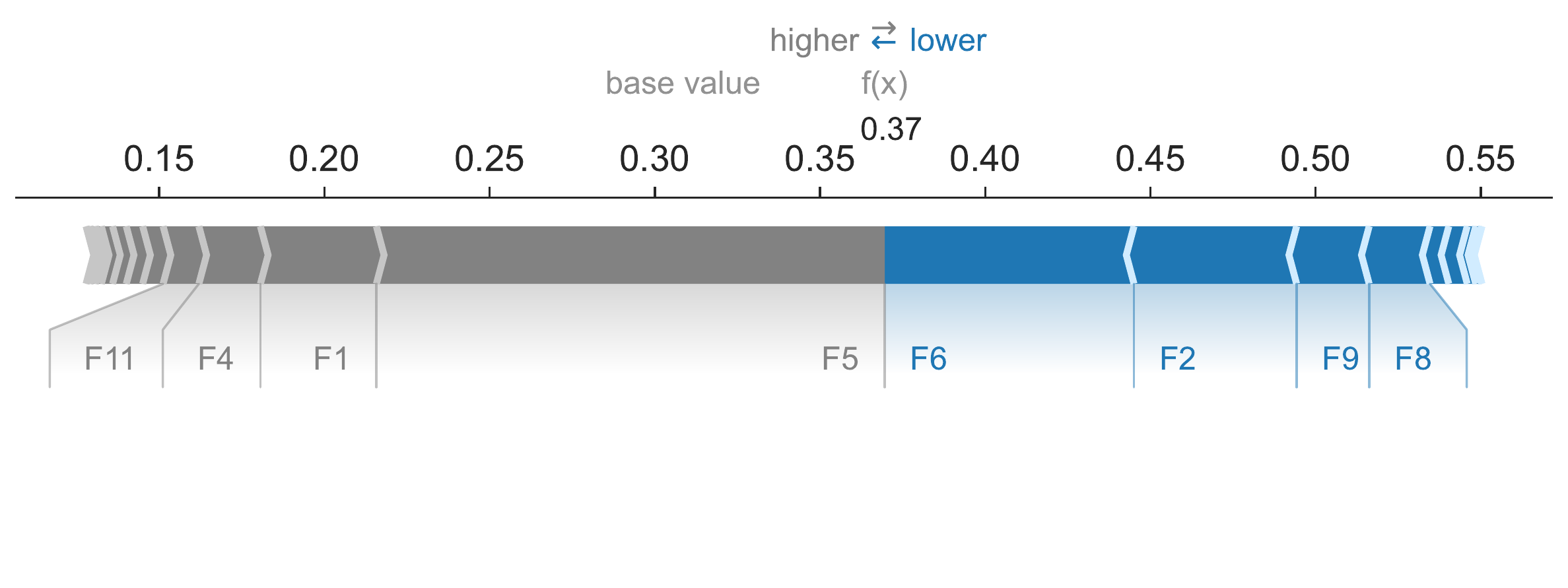}
	\caption{negative instance}
	\end{subfigure}
	
	\captionsetup{justification=justified}
	\caption{Force plot depicting feature attribution for randomly selected positive and negative instances of the German Credit dataset. The results show that the protected features do not contribute notably to the predictions.}
\label{fig:force-german}
\end{figure*}

Figure \ref{fig:force-german} shows the feature importance computed by SHAP for randomly selected positive and negative instances. The model's prediction for the positive instance is 0.98 in terms of the probability of obtaining a good credit risk assessment. The width of the bars corresponds to the magnitude of the feature attribution. For example, the checking account (F1) contributes positively to increasing the probability with a value of 0.11, compared to the average prediction of the dataset for the positive class (0.70). In the same way, the features credit amount (F5) and savings account (F6) contribute 0.04 each, while other features such as other installment (F14), duration (F2), \textit{gender} (F9), and housing (F15) have smaller positive contributions. The value of purpose (F4) reduces the probability of getting a positive outcome for this instance.

In contrast, the model's prediction for the randomly selected negative instance is 0.30, while the prediction for this instance is 0.37. In this case, the difference with the base value is not large, resulting from the combination of positive and negative contributions of several features. For example, credit amount (F5) contributes to increasing the probability, while savings account (F6) decreases the probability by a magnitude of 0.08. Other features contribute negatively and positively to the instance prediction.

\begin{figure*}[!htb]
\center

    \begin{subfigure}{0.49\textwidth}
	\center
	\includegraphics[width=\textwidth]{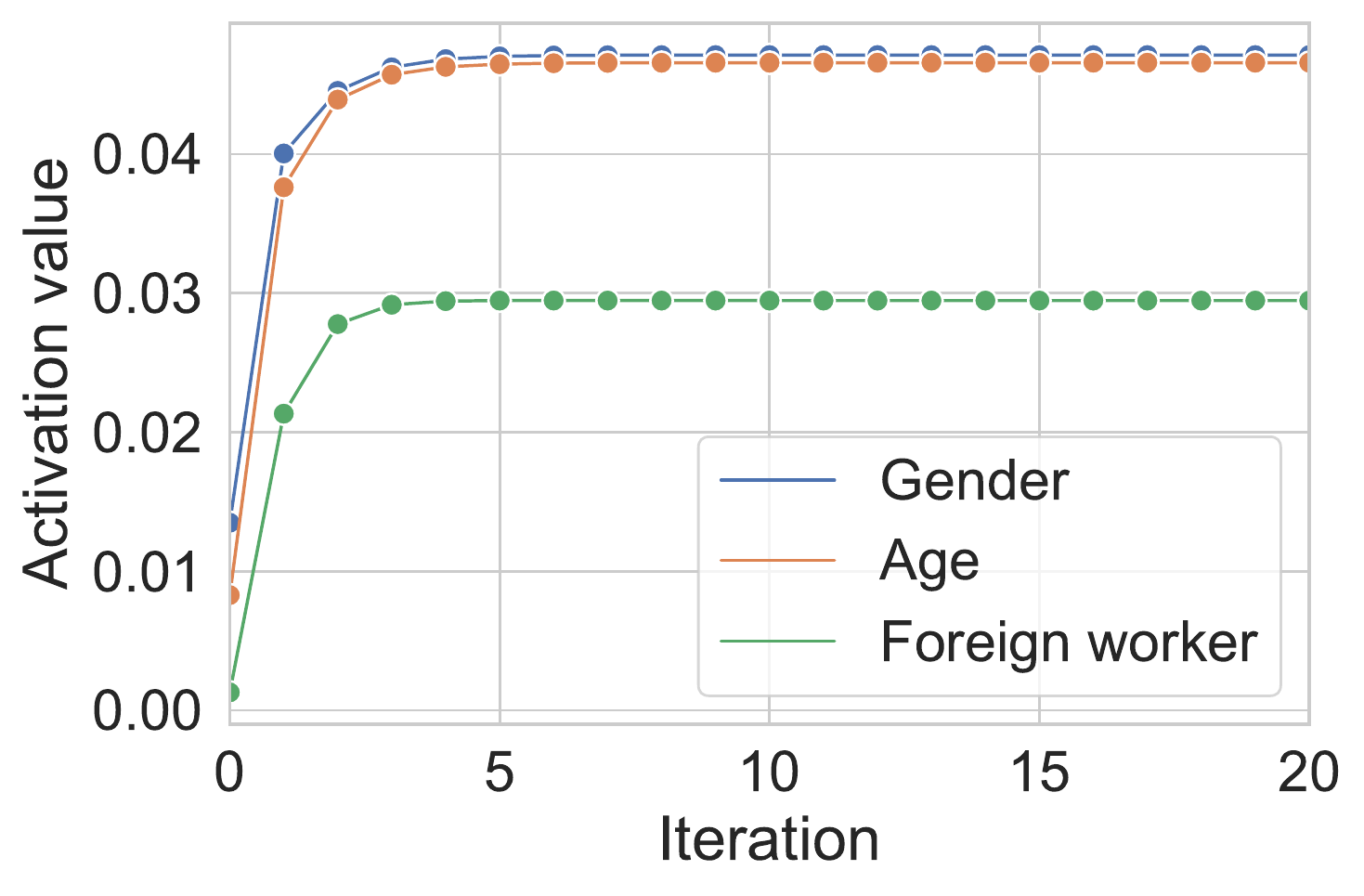}
	\caption{$\phi=0.2$}
	\end{subfigure}
	\begin{subfigure}{0.49\textwidth}
	\center
	\includegraphics[width=\textwidth]{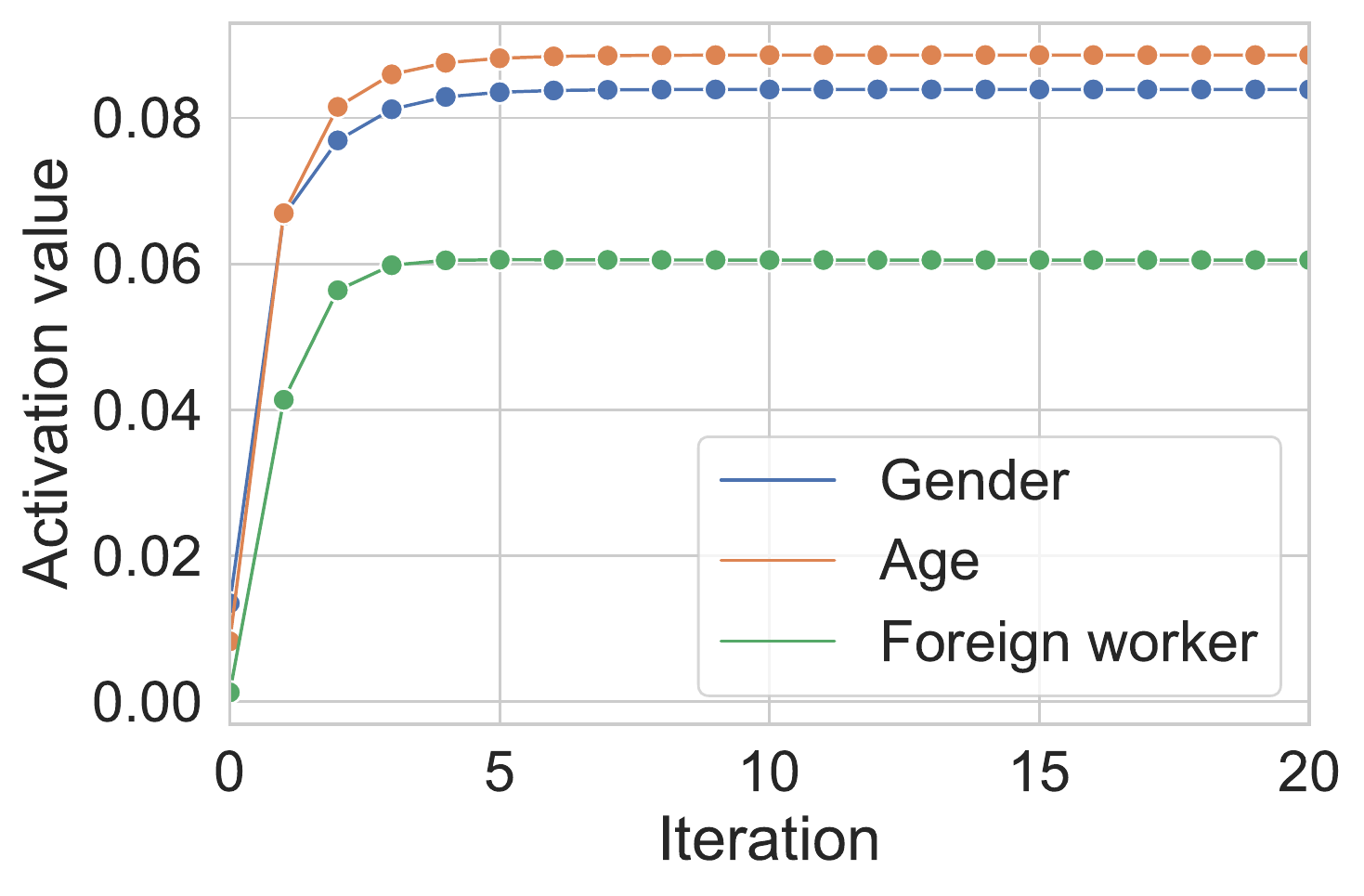}
	\caption{$\phi=0.4$}
	\end{subfigure}
	
	\begin{subfigure}{0.49\textwidth}
	\center
	\includegraphics[width=\textwidth]{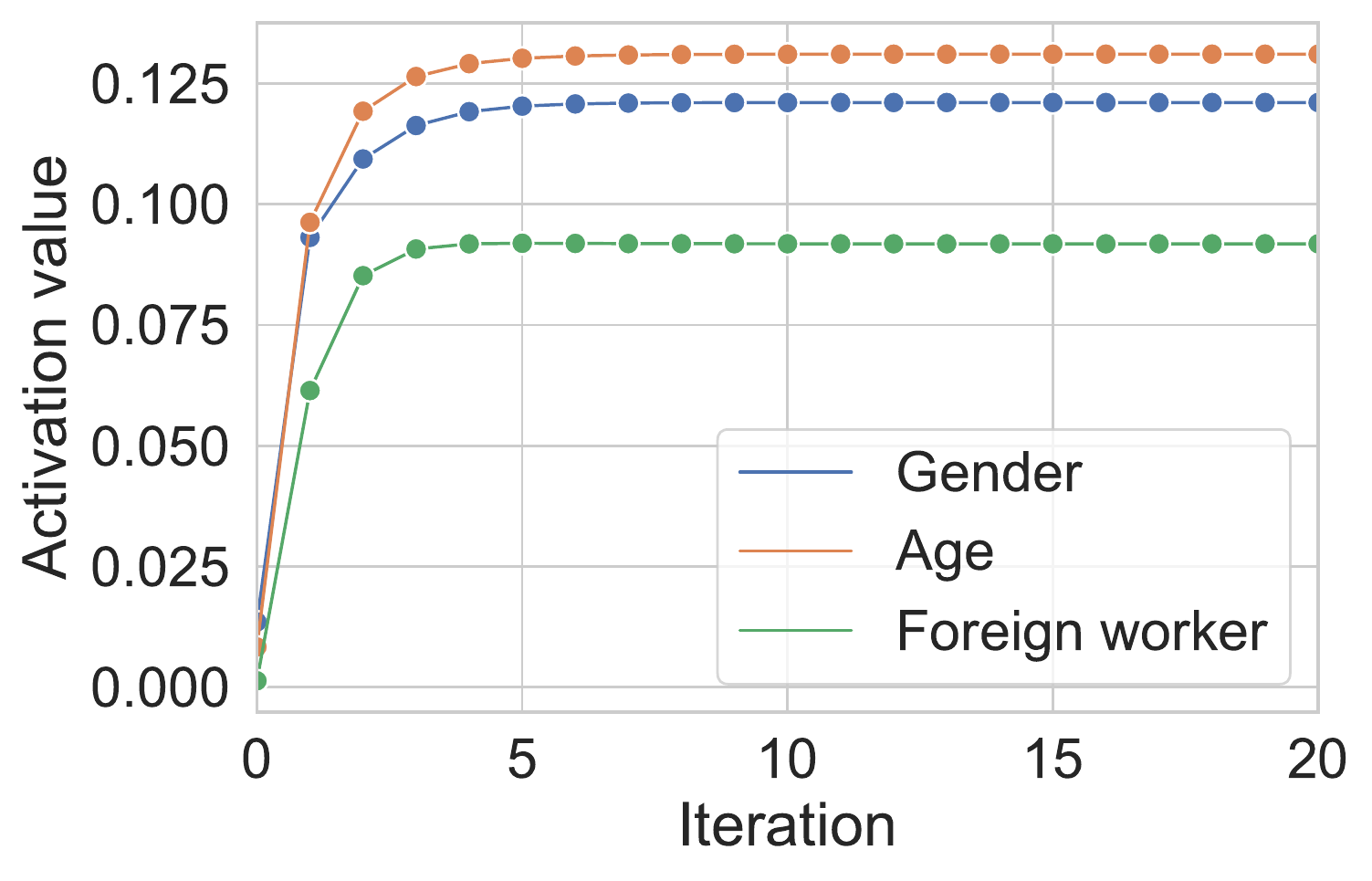}
	\caption{$\phi=0.6$}
	\end{subfigure}
	\begin{subfigure}{0.49\textwidth}
	\center
	\includegraphics[width=\textwidth]{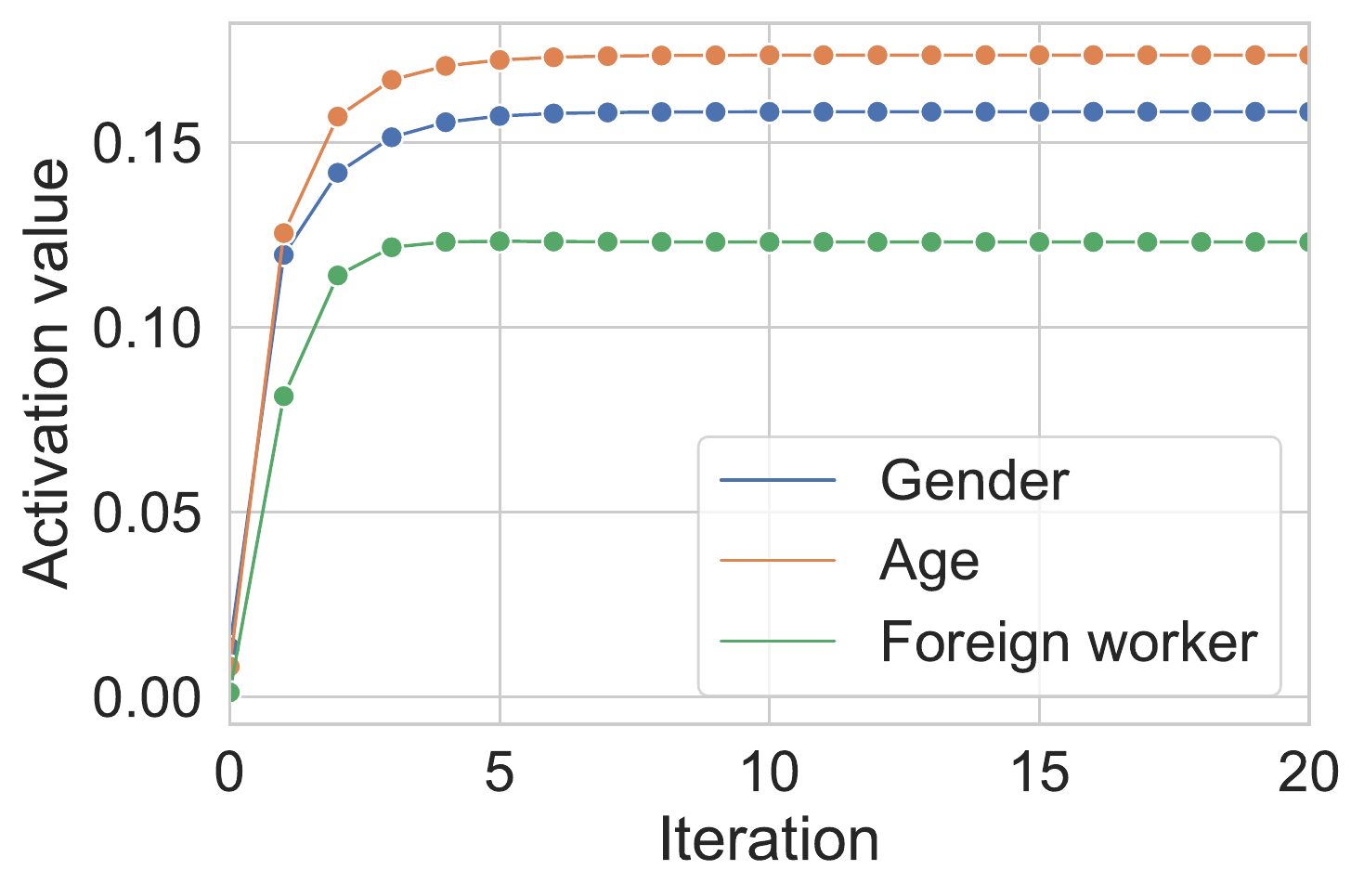}
	\caption{$\phi=0.8$}
	\end{subfigure}
	
	\captionsetup{justification=justified}
	\caption{Activation values of neurons denoting protected features for a positive instance in the German Credit dataset for different $\phi$ values. Although the neurons associated with the protected features are initialized with very small values, we can observe an increase in their activation values.}
\label{fig:fcm_good}
\end{figure*}

Next, we use the SHAP values as an initialization vector in our FCM model for studying the implicit bias starting from the random instances above. For the first instance, Figure \ref{fig:fcm_good} depicts the activation values of the protected features (\textit{gender}, \textit{age}, and \textit{foreign worker}). Although the initial activation values of neurons denoting these features are rather low (0.01 for \textit{gender} and \textit{age}, and 0 for \textit{foreign worker}), we can see a clear increase in their activation after a few iterations. This pattern is consistent across different values of $\phi$, obtaining higher values as $\phi$ increases. This increment comes from the interaction with other features, suggesting the presence of implicit bias in the dataset.

\begin{figure*}[!htb]
\center

    \begin{subfigure}{0.49\textwidth}
	\center
	\includegraphics[width=\textwidth]{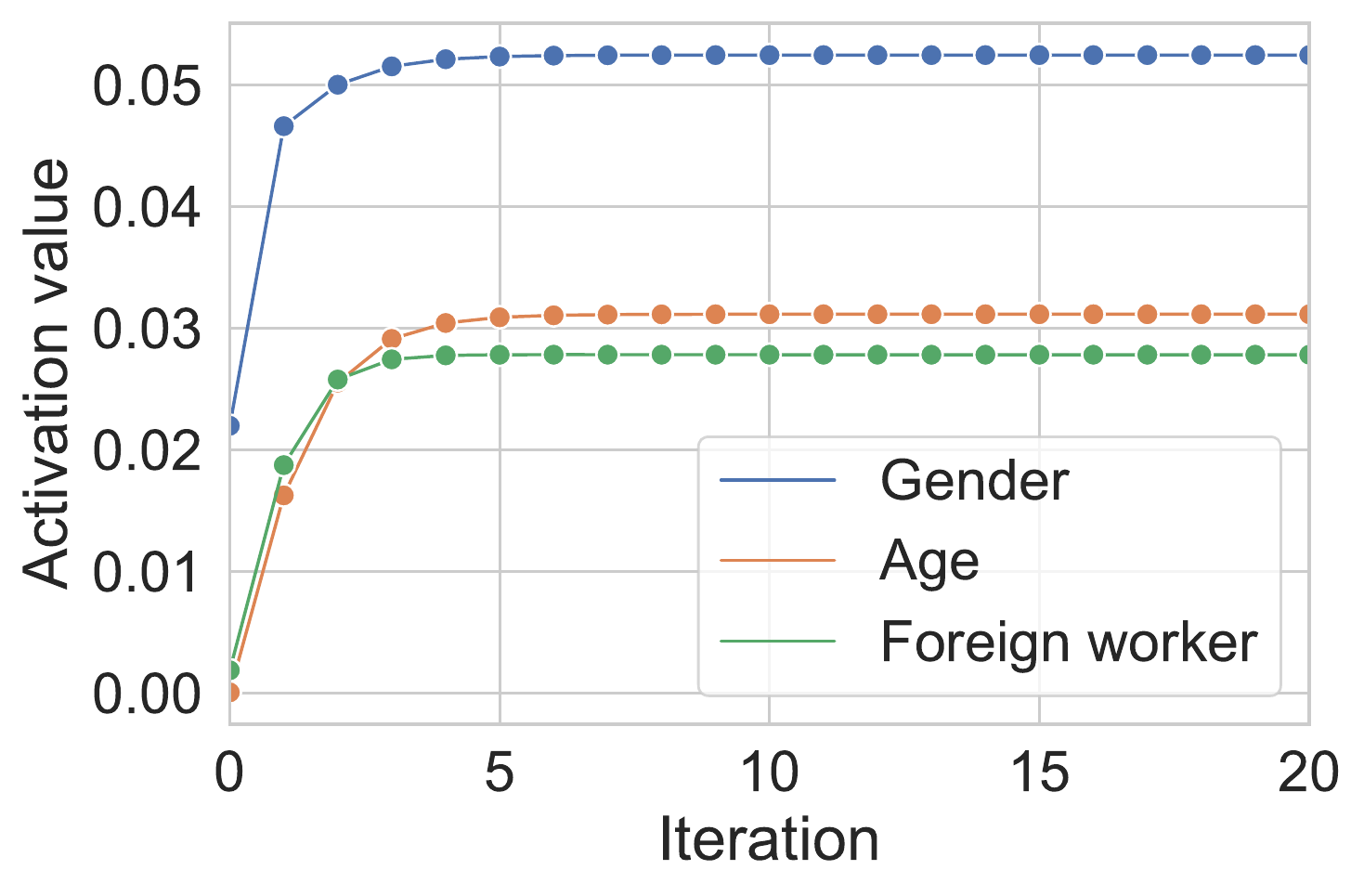}
	\caption{$\phi=0.2$}
	\end{subfigure}
	\begin{subfigure}{0.49\textwidth}
	\center
	\includegraphics[width=\textwidth]{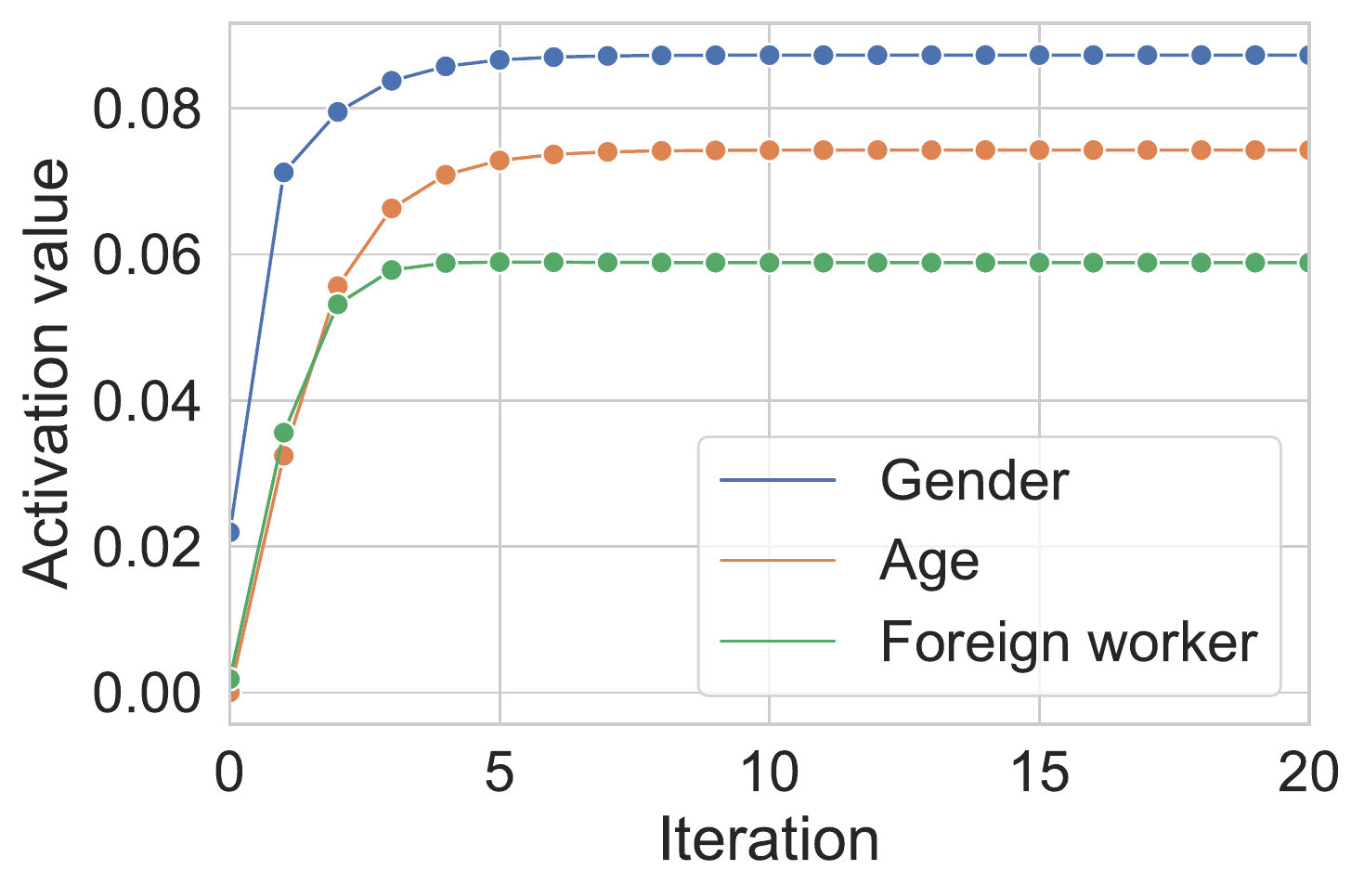}
	\caption{$\phi=0.4$}
	\end{subfigure}
	
	\begin{subfigure}{0.49\textwidth}
	\center
	\includegraphics[width=\textwidth]{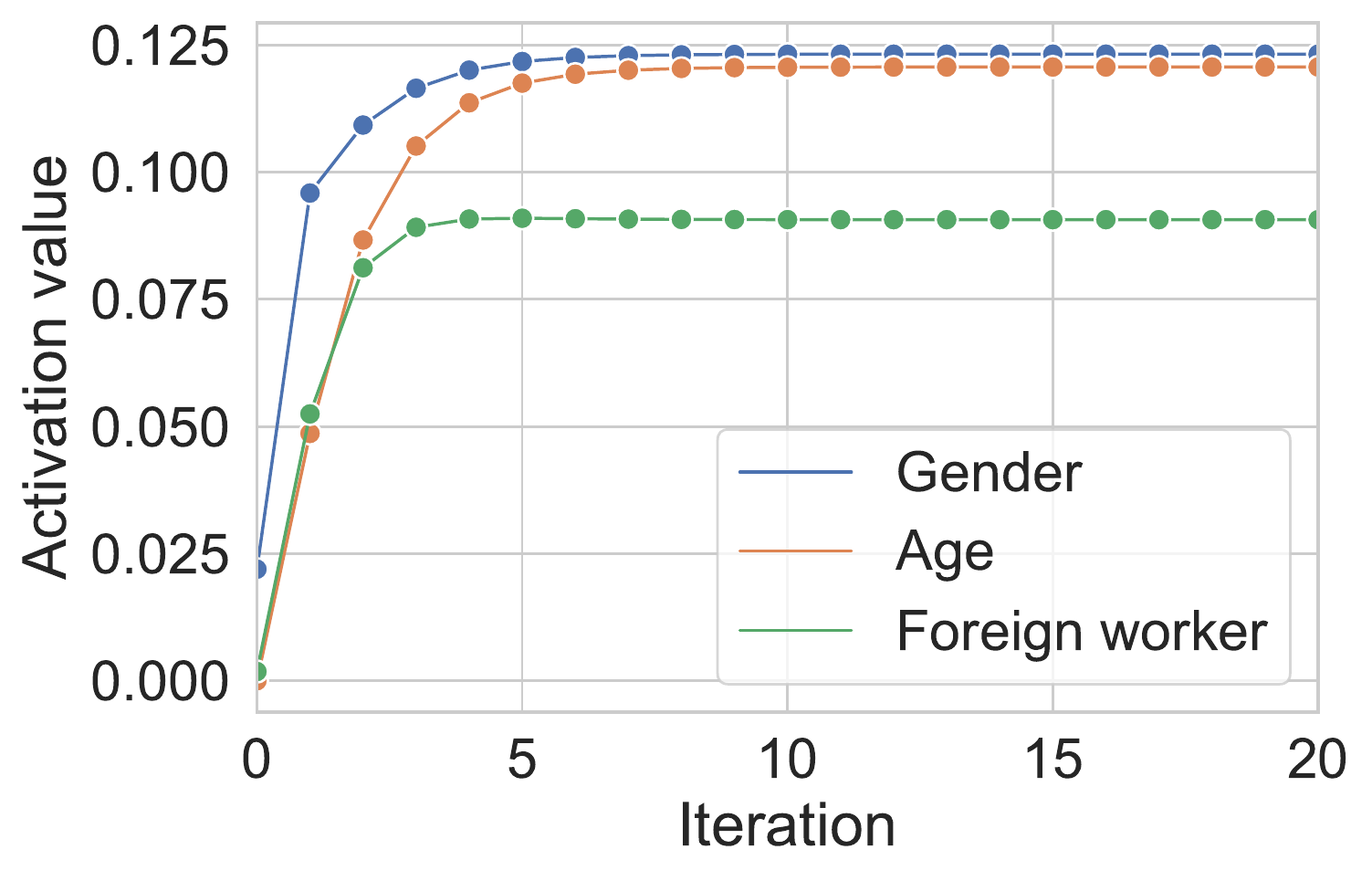}
	\caption{$\phi=0.6$}
	\end{subfigure}
	\begin{subfigure}{0.49\textwidth}
	\center
	\includegraphics[width=\textwidth]{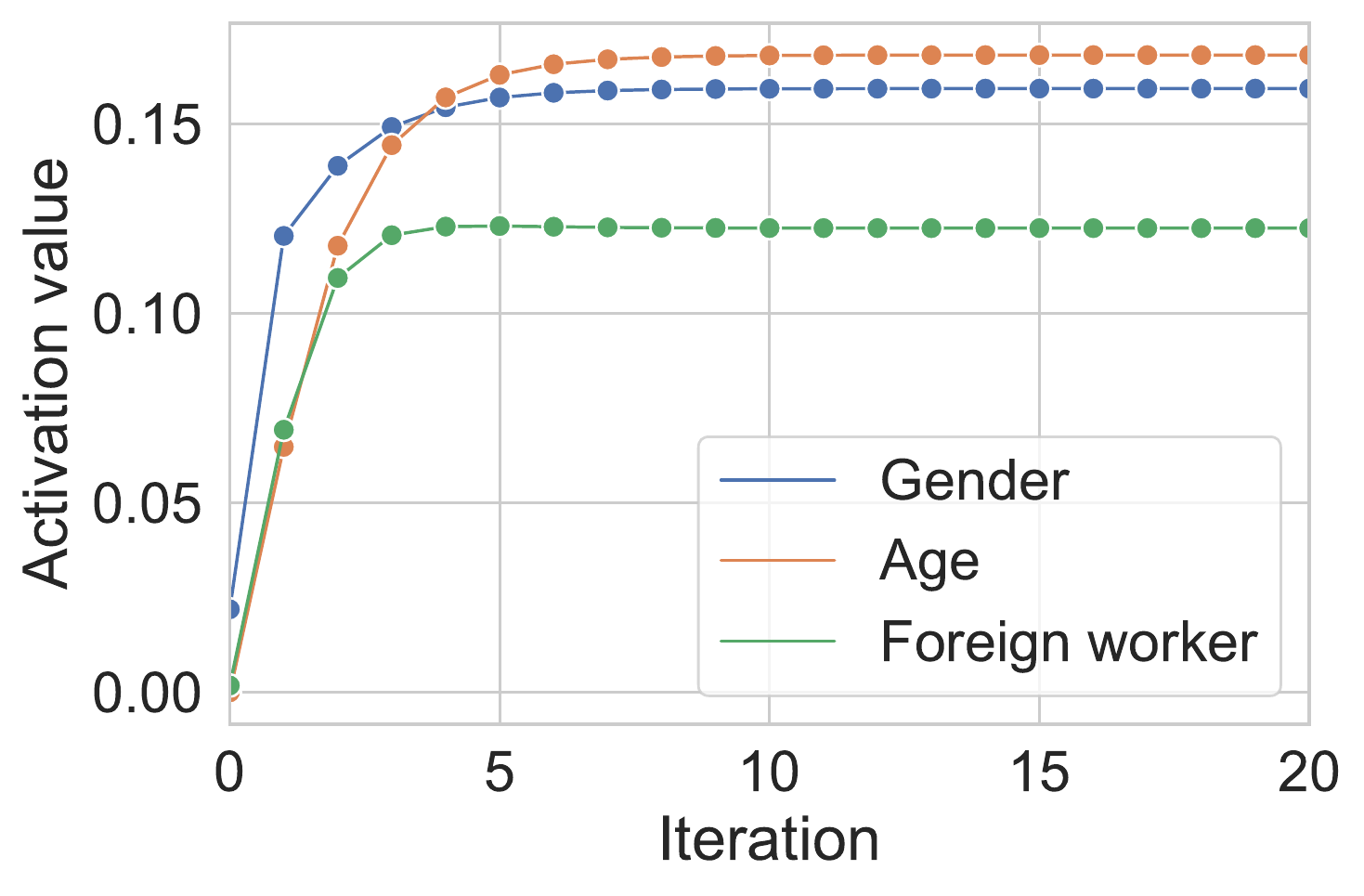}
	\caption{$\phi=0.8$}
	\end{subfigure}
	
	\captionsetup{justification=justified}
	\caption{Activation values of neurons denoting protected features for a negative instance in the German Credit dataset for different $\phi$ values. Although the neurons associated with the protected features are initialized with very small values, we can observe an increase in their activation values.}
\label{fig:fcm_bad}
\end{figure*}

We repeat the analysis using the SHAP values corresponding to the negative instance as the initialization vector in our FCM  model. Figure \ref{fig:fcm_bad} plots the activation values of the protected features \textit{gender}, \textit{age}, and \textit{foreign worker}. The initial SHAP values for the protected features are small, with \textit{gender} (F9) having an activation value of 0.02 and \textit{age} (F13) and \textit{foreign worker} (F20) having no direct contribution. However, after a few iterations, we can observe an increase in the values of the three protected features, with \textit{gender} (F9) and \textit{age} (F13) as the most excited neurons. This pattern is also consistent across different $\phi$ values, where the higher the $\phi$ value, the more pronounced the increment. Recall that the smaller the $\phi$ parameter, the more linear the model, therefore the activation values are more similar to the initial feature importance. 

\begin{figure*}[!htbp]
\center

    \begin{subfigure}{0.8\textwidth}
	\center
	\includegraphics[width=\textwidth]{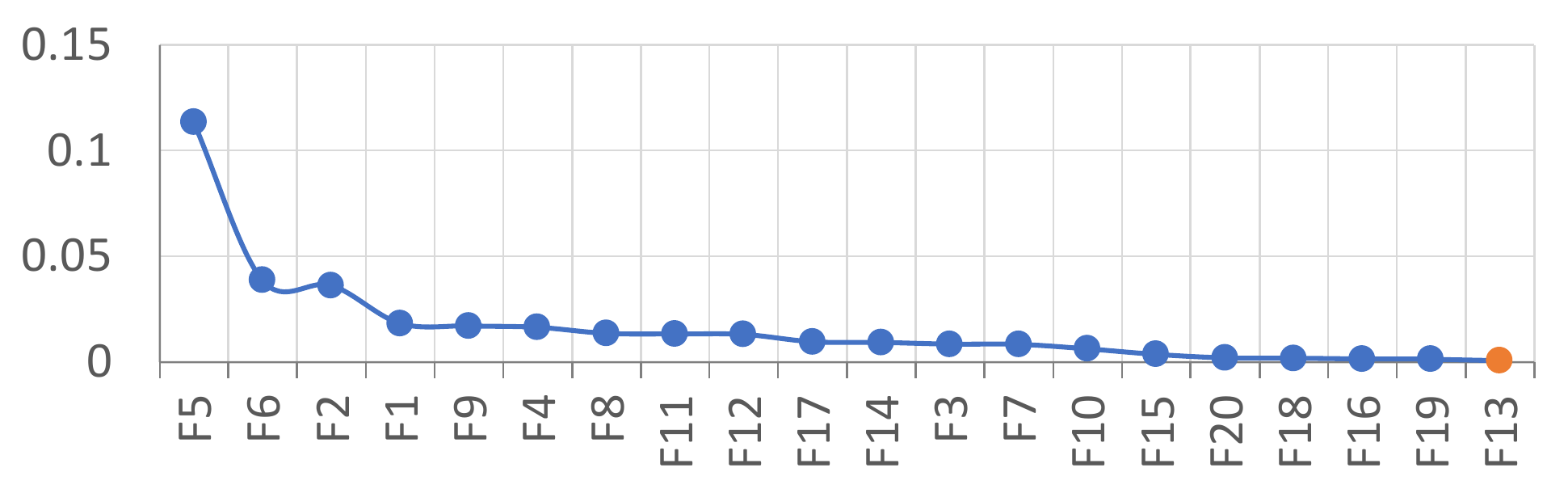}
	\caption{SHAP values}
	\end{subfigure}
	\begin{subfigure}{0.8\textwidth}
	\center
	\includegraphics[width=\textwidth]{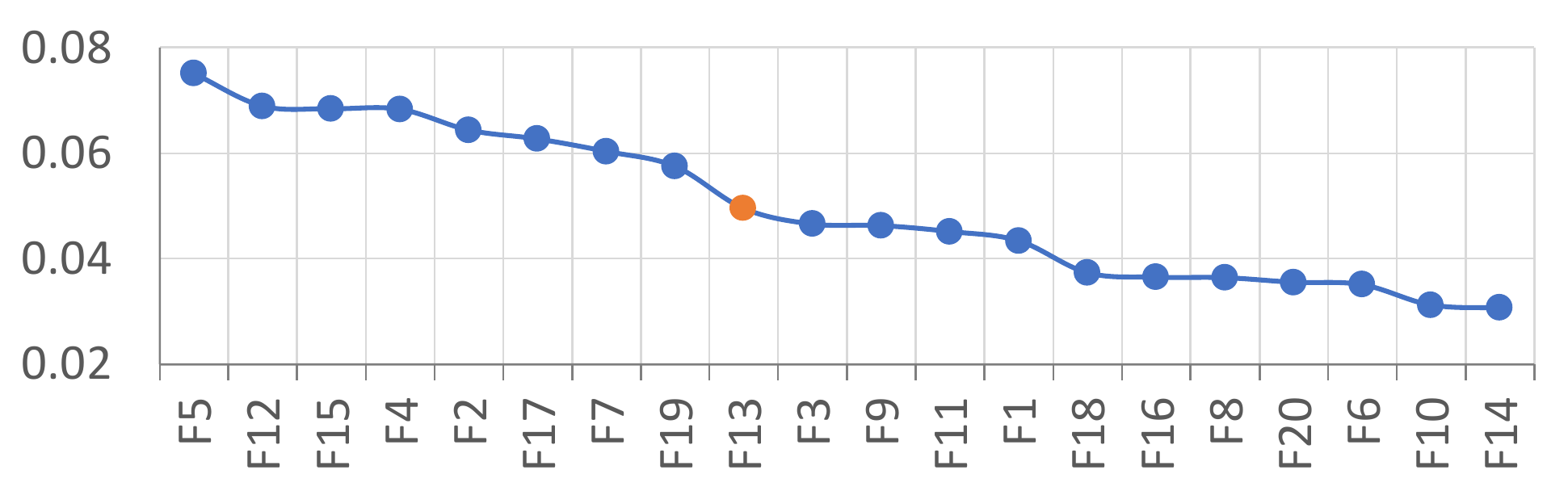}
	\caption{FCM values}
	\end{subfigure}
	
	\captionsetup{justification=justified}
	\caption{Absolute SHAP values and bias scores computed by the FCM model for the randomly selected negative instance of the German Credit dataset. In this case, the protected feature \textit{age} (F13) is the least relevant according to SHAP, but this feature does involve a moderate amount of bias.}
\label{fig:comparison)}
\end{figure*}

The reader could argue that Figures \ref{fig:fcm_good} and \ref{fig:fcm_bad} do not allow assessing whether the protected features rank comparably w.r.t. the SHAP values and the outputs of the FCM model. In other words, the activation values of neurons representing protected features might be significantly smaller than those denoting unprotected features. Figure \ref{fig:comparison)} displays the aggregated SHAP values and the normalized activation values produced by the FCM model ($\phi=0.8$) in the last iteration for the negative instance. This figure shows that \textit{age} (F13) is deemed the least relevant feature when classifying the instance according to the SHAP values. However, the amount of bias captured by the FCM model against \textit{age} is quite significant. At the same time, the protected feature \textit{gender} (F9) was recognized by SHAP and the FCM model as important. Finally, both the SHAP values and the FCM model agree that \textit{foreign worker} (F20) does not seem to play a relevant role when classifying the instance.

The simulations provide evidence of implicit bias even when protected features are not deemed explicitly important according to SHAP values. This happens because features are not independent, and as such, unprotected features can partially encode the information of protected ones.

The final point to be discussed is whether the conclusions concerning implicit bias change if protected numeric features are categorically encoded. In the fairness literature, numeric features are often associated with protected groups such as females or young people. While the results in Figures \ref{fig:fcm_good} and \ref{fig:fcm_bad} report more biased against \textit{age} than \textit{gender}, the opposite is concluded in \cite{Napoles2022a} and \cite{Napoles2022c} in which protected features are analyzed as a whole. The cause of this difference is that in the approach presented in this paper, the fuzzy $c$-means algorithm automatically detects such groups. This remark agrees with the results presented in \cite{Napoles2022a}, where the authors showed how analyzing bias at a group level leads to different conclusions, even using the same model and metrics. 

\subsection{Adult dataset}

The second case study concerns the Adult dataset~\cite{Kohavi1996ScalingUT}. For this dataset, the pre-processing step only involved the normalization of numeric features. Table \ref{table:association_adult} shows the association values between all features and protected features \textit{race} (F9) and \textit{sex} (F10) in Adult dataset along with the global SHAP values per feature. We observe that some unprotected features have rather strong associations with protected ones: native-country (F14) is strongly associated with \textit{race} (F9), and the protected feature \textit{sex} (F10) is strongly associated with marital-status (F6), occupation (F7) and relationship (F8). According to the global SHAP values, marital-status (F6) and relationship (F8) are key features when making the decision, while both protected features are deemed relatively irrelevant. 

\begin{table}[!htb]
\centering
\caption{Association values between protected and unprotected features in the Adult dataset. Global SHAP values provide information about feature importance using a Random Forest as a classifier.\\}
\label{table:association_adult}
\begin{tabular}{l|l|c|c|c}
\hline
\multicolumn{1}{c|}{\multirow{2}{*}{ID}} & \multicolumn{1}{c|}{\multirow{2}{*}{Features}} & \multicolumn{2}{c|}{Associates with} & \multirow{2}{*}{SHAP}  \\ 
\cline{3-4}
\multicolumn{1}{c|}{} & \multicolumn{1}{c|}{} & Race & Sex &\\ 
\hline

F1  &             Age &  0.04 &   0.07 &  0.039 \\
F2  &       Workclass &  0.06 &   0.15 &  0.011 \\
F3  &          Fnlwgt &  0.11 &   0.04 &  0.010 \\
F4  &       Education &  0.07 &    0.1 &  0.026 \\
F5  &   Education-num &  0.07 &   0.09 &  0.037 \\
F6  &  Marital-status &  0.08 &   0.46 &  0.066 \\
F7  &      Occupation &  0.08 &   0.42 &  0.040 \\
F8  &    Relationship &   0.1 &   0.65 &  0.055 \\
\cellcolor{Gray}F9  & \cellcolor{Gray}Race &   1.0 &   0.12 &  0.004 \\
\cellcolor{Gray}F10 & \cellcolor{Gray}Sex &  0.12 &    1.0 &  0.014 \\
F11 &    Capital-gain &  0.02 &   0.03 &  0.046 \\
F12 &    Capital-loss &  0.03 &   0.05 &  0.012 \\
F13 &  Hours-per-week &  0.06 &   0.26 &  0.026 \\
F14 &  Native-country &  0.41 &   0.07 &  0.003 \\

\hline
\end{tabular}

\end{table}

\begin{figure*}[!htb]
\center
    \begin{subfigure}{0.9\textwidth}
	\center
	\includegraphics[width=\textwidth, trim=0 2.2cm 0 0,clip]{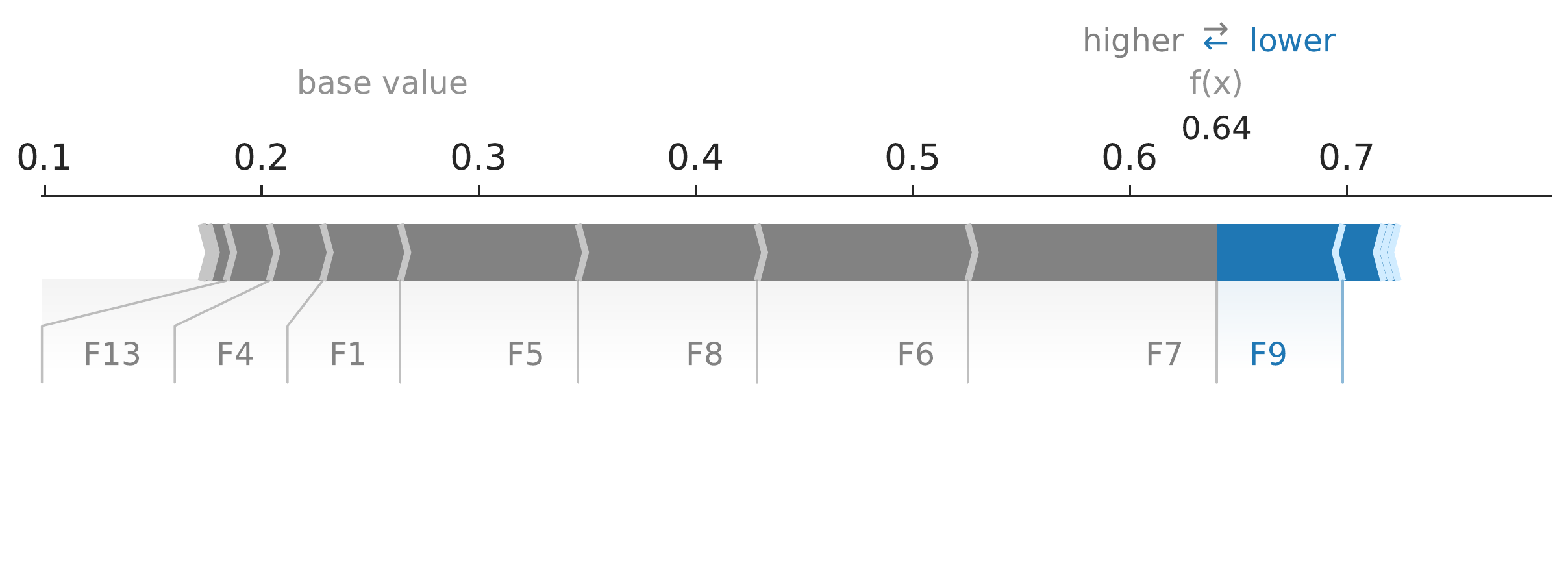}
	\caption{positive instance}
	\end{subfigure}
	\begin{subfigure}{0.9\textwidth}
	\center
	\includegraphics[width=\textwidth, trim=0 2.2cm 0 -0.5cm,clip]{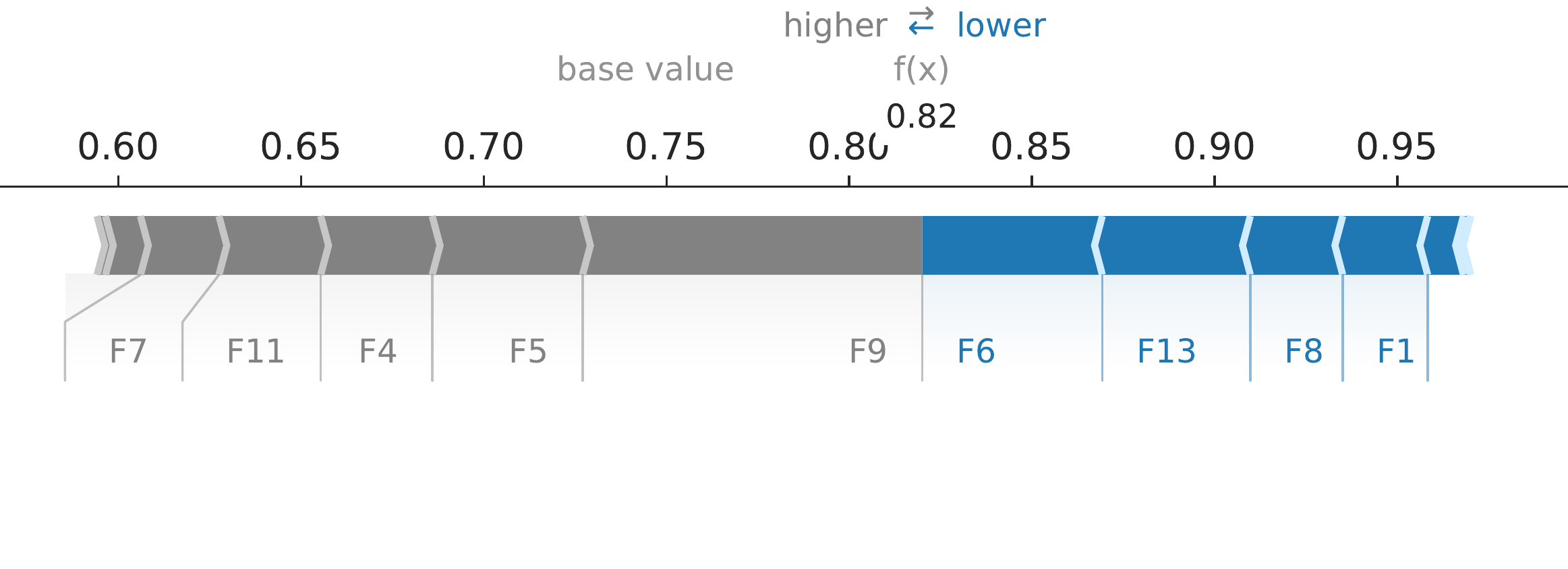}
	\caption{negative instance}
	\end{subfigure}
	
	\captionsetup{justification=justified}
	\caption{Force plot depicting feature attribution for randomly selected positive and negative instances of the Adult dataset. The results show that the protected feature F9 contributes to the predictions.}
\label{fig:force-adult}
\end{figure*}

Figure \ref{fig:force-adult} displays the feature importance computed by SHAP for randomly selected positive and negative instances, which are selected from the test set after verifying they have been correctly classified.

Figure \ref{fig:force-adult} shows that the model's prediction for the positive class is 0.26, while the prediction for this instance is 0.64. Features occupation (F7), marital-status (F6), and relationship (F8) contribute positively to increasing the probability. In contrast, the \textit{race} (F9) feature reduces the probability of getting a positive outcome for this instance by 0.06.

The overall dataset prediction for the negative class is 0.74, while the prediction for the randomly selected negative instance is 0.82. Feature \textit{race} (F9) has the largest positive contribution (0.09) followed by education-num (F5) and education (F4). Features marital-status (F6), hours-per-week (F13) and relationship (F8) reduce the probability of getting a positive outcome for this instance.

\begin{figure*}[!htb]
\center

    \begin{subfigure}{0.49\textwidth}
	\center
	\includegraphics[width=\textwidth]{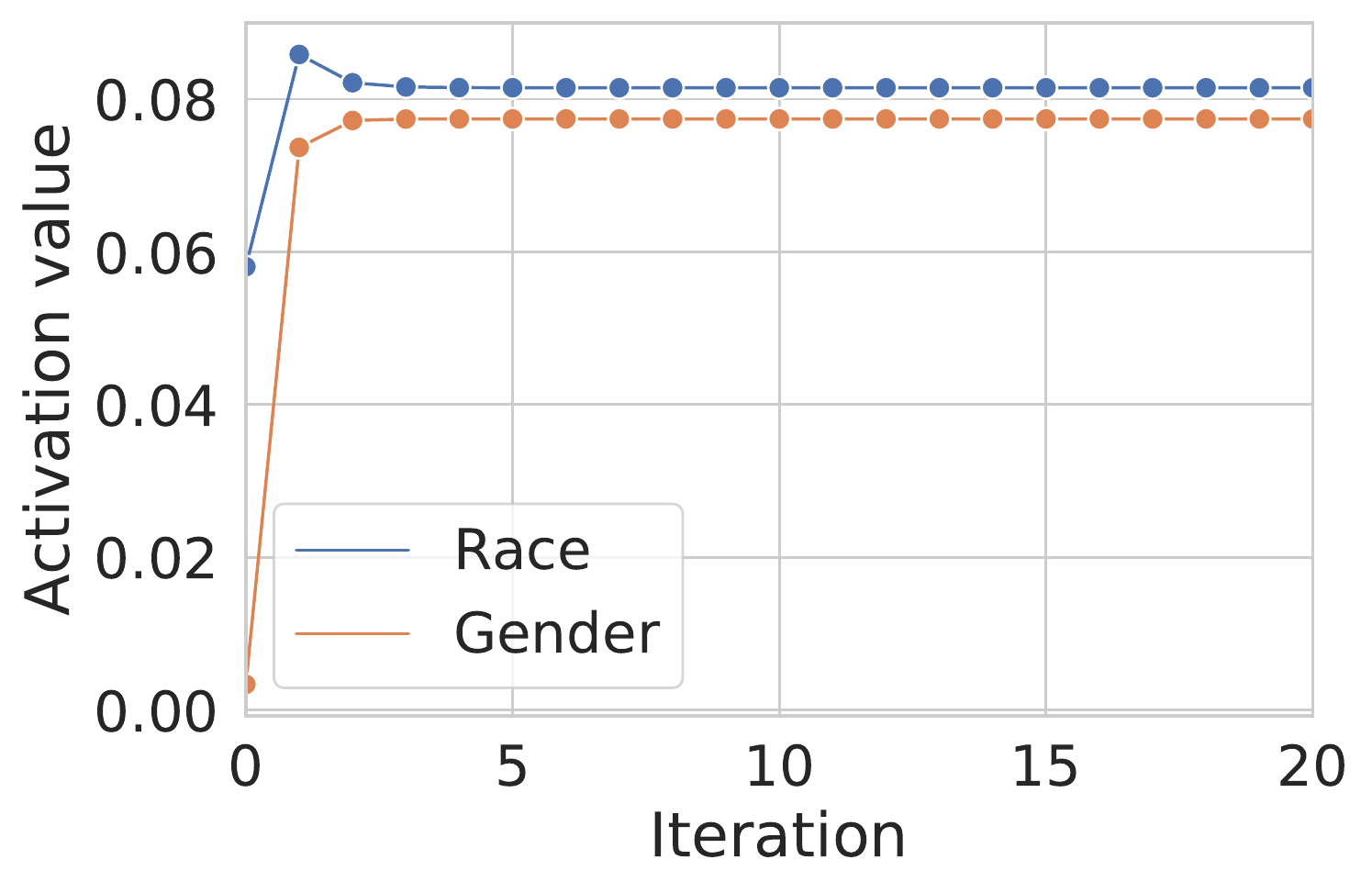}
	\caption{$\phi=0.2$}
	\end{subfigure}
	\begin{subfigure}{0.49\textwidth}
	\center
	\includegraphics[width=\textwidth]{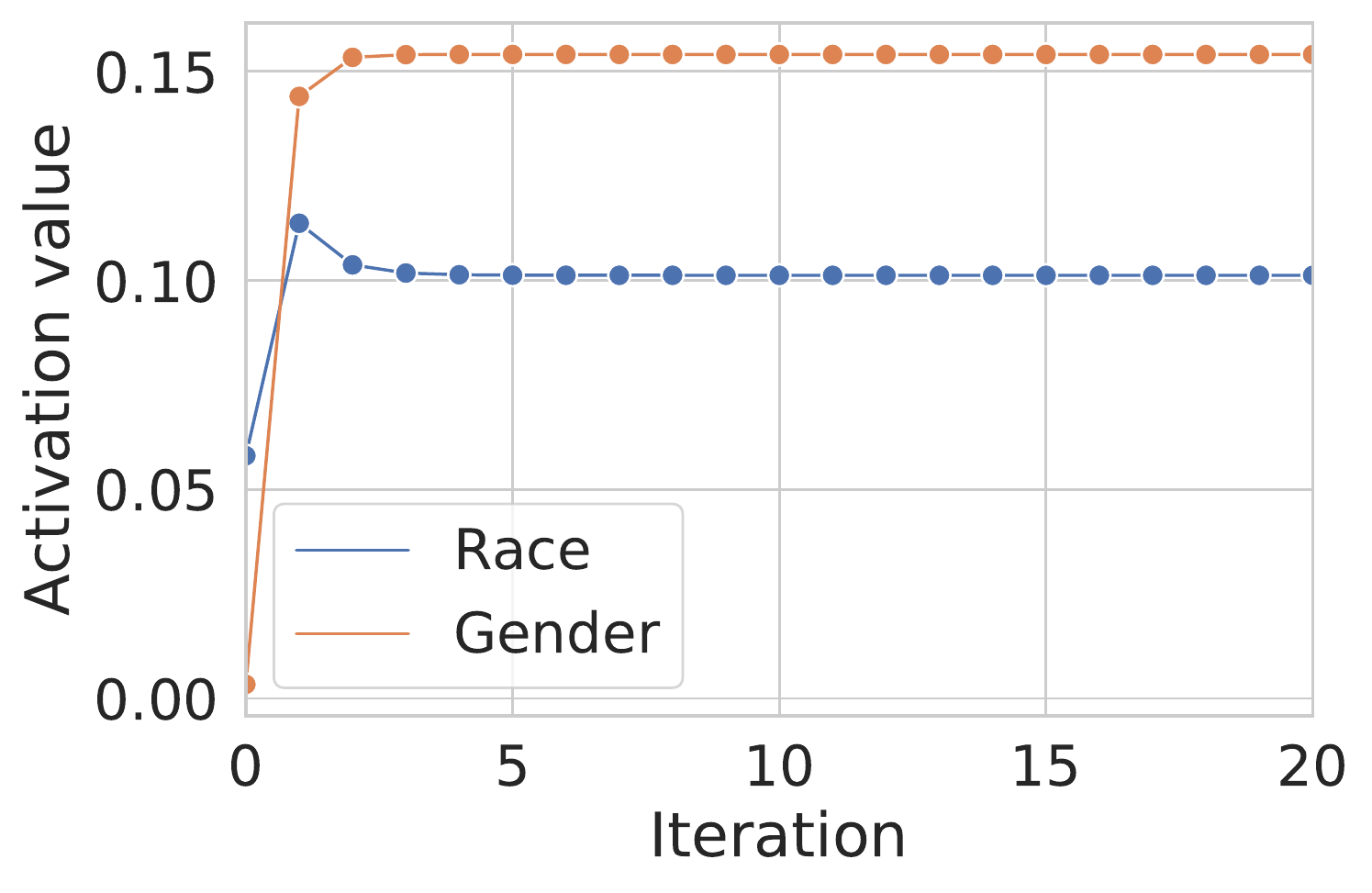}
	\caption{$\phi=0.4$}
	\end{subfigure}
	\begin{subfigure}{0.49\textwidth}
	\center
	\includegraphics[width=\textwidth]{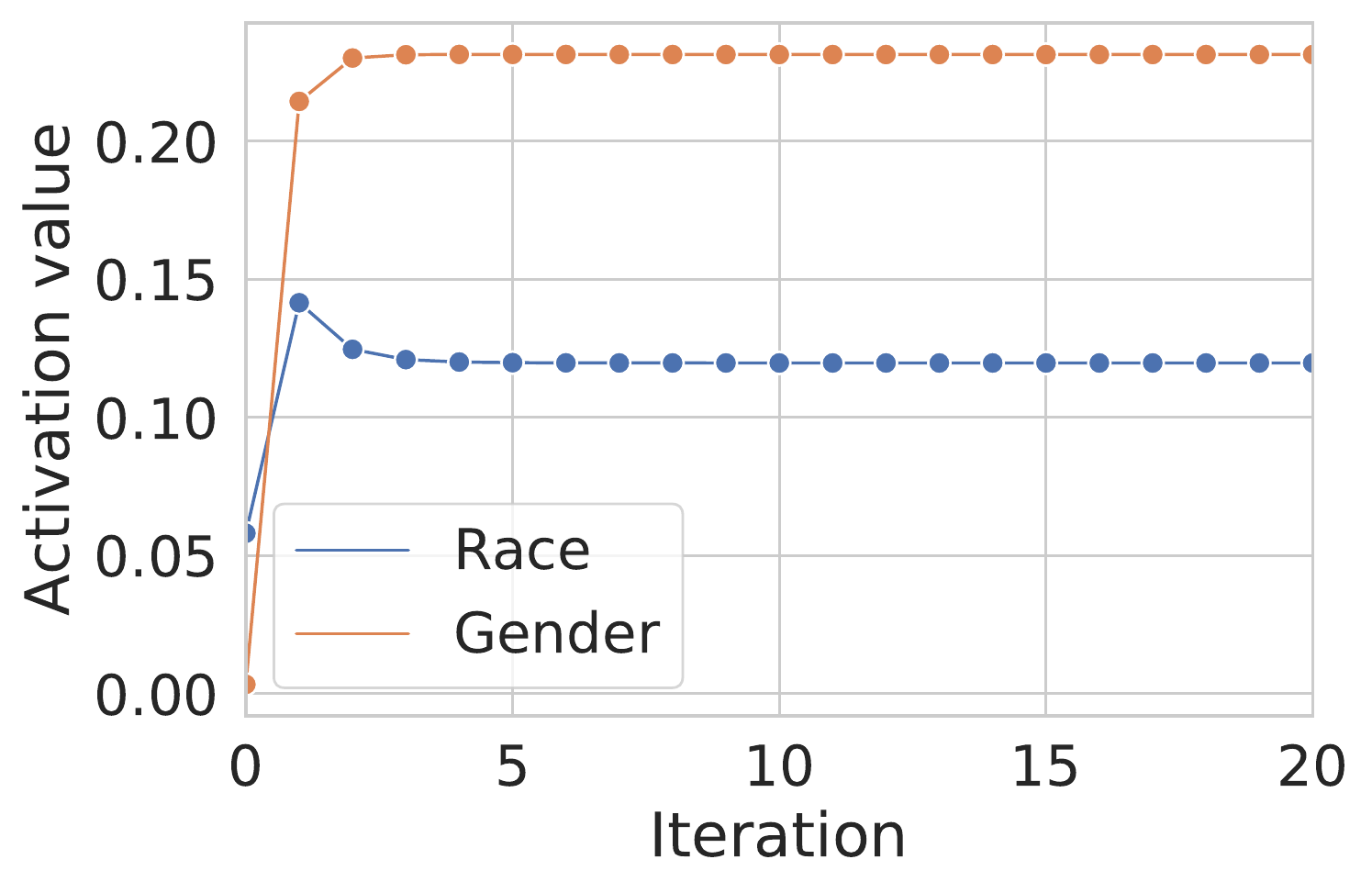}
	\caption{$\phi=0.6$}
	\end{subfigure}
	\begin{subfigure}{0.49\textwidth}
	\center
	\includegraphics[width=\textwidth]{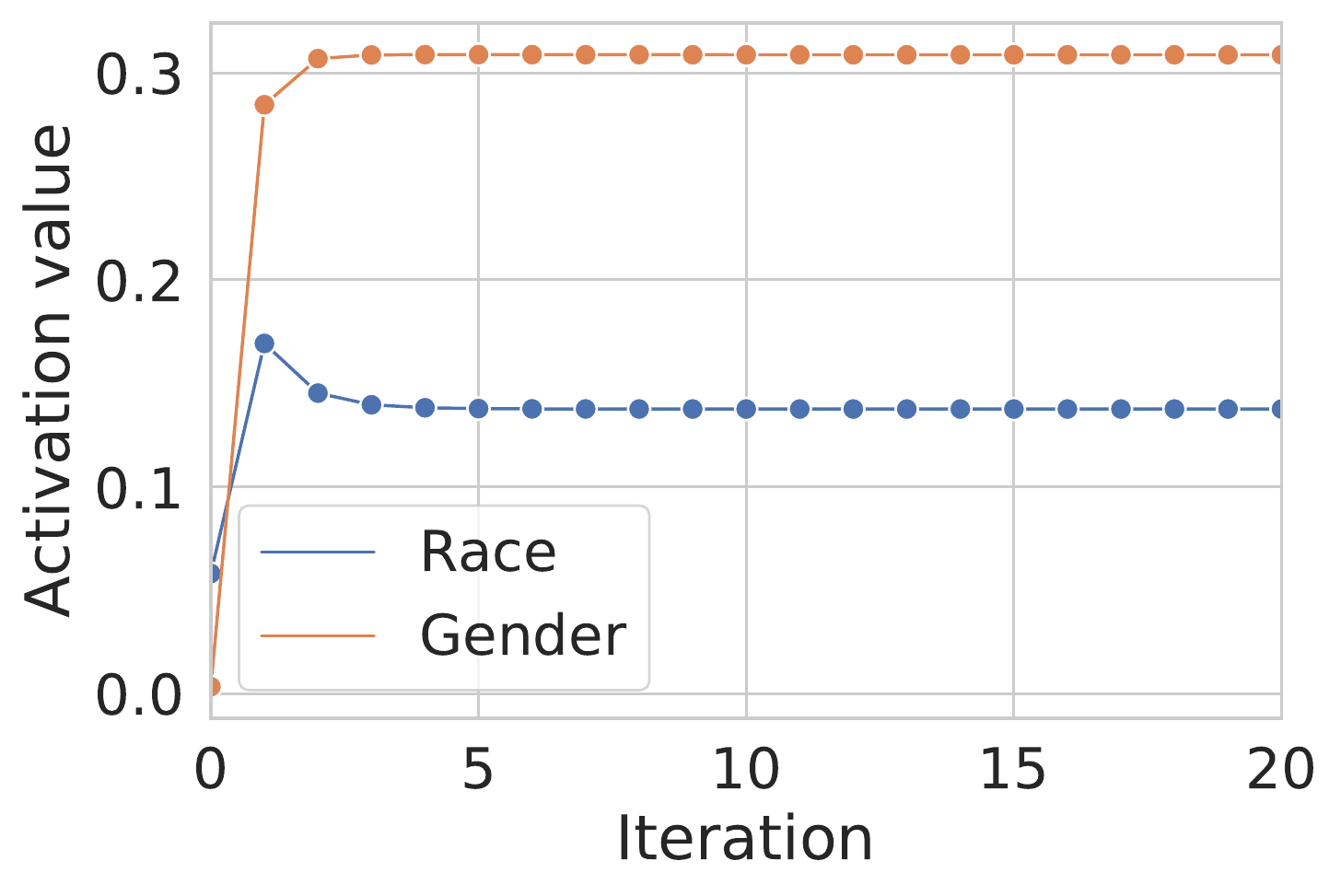}
	\caption{$\phi=0.8$}
	\end{subfigure}
	
	\captionsetup{justification=justified}
	\caption{Activation values of neurons denoting protected features for a positive instance in the Adult dataset for different $\phi$ values.}
\label{fig:fcm_good_adult}
\end{figure*}

Next, we use the SHAP values as an initialization vector in our FCM model for studying the implicit bias starting from the two randomly selected instances. Figure \ref{fig:fcm_good_adult} depicts the activation values of the protected features (\textit{race} and \textit{sex}). Although the initial activation values of neurons denoting \textit{sex} (F10) start close to zero, we can see a clear increase in their activation values after a few iterations for all levels of $\phi$, obtaining higher values as $\phi$ increases reaching 0.3 at $\phi=0.8$. Comparatively, \textit{race} (F9) does not deviate much from its initial activation value (starts at 0.05 and reaches 0.14 at $\phi=0.8$).

\begin{figure*}[!htb]
\center

    \begin{subfigure}{0.49\textwidth}
	\center
	\includegraphics[width=\textwidth]{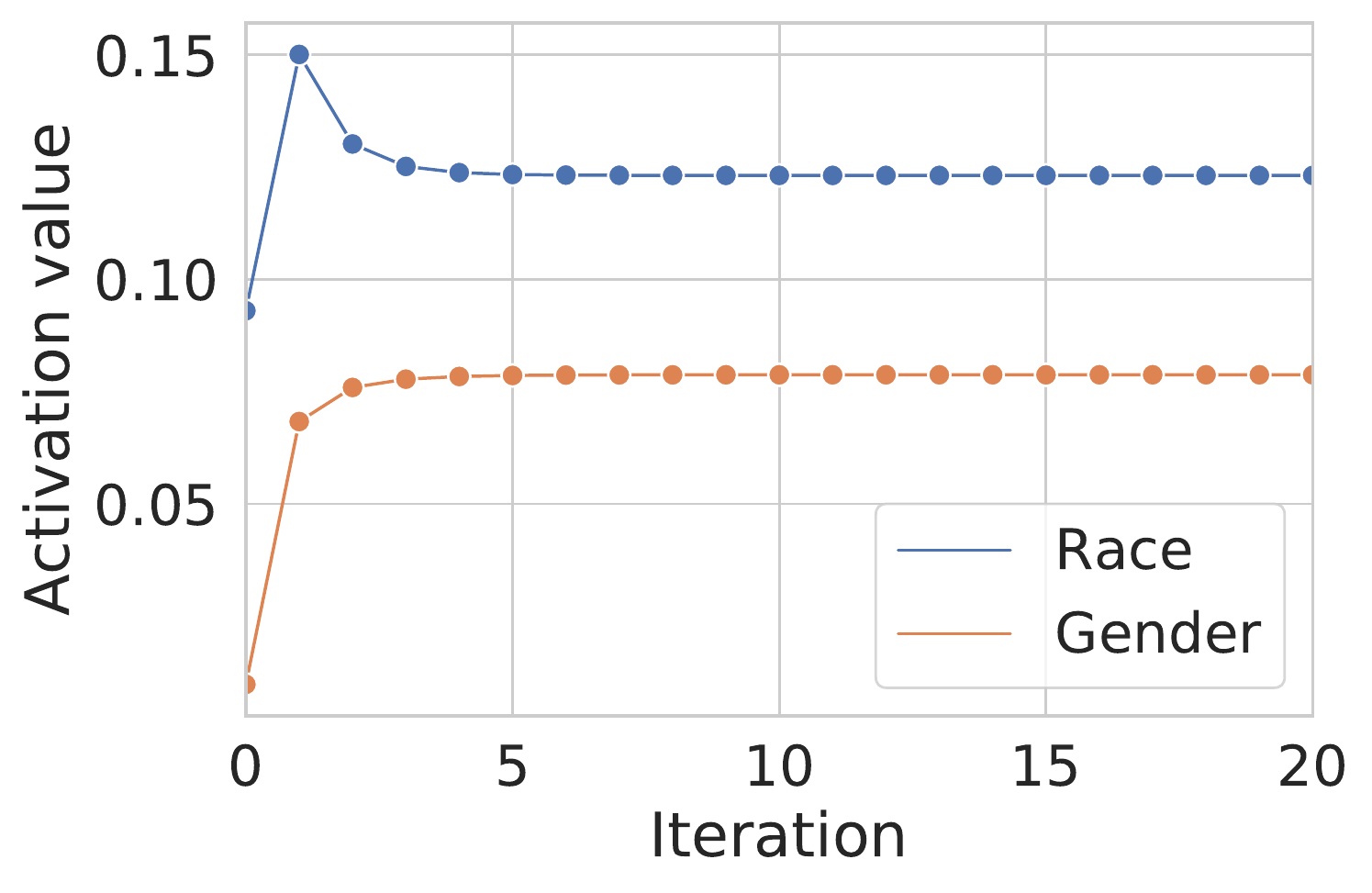}
	\caption{$\phi=0.2$}
	\end{subfigure}
	\begin{subfigure}{0.49\textwidth}
	\center
	\includegraphics[width=\textwidth]{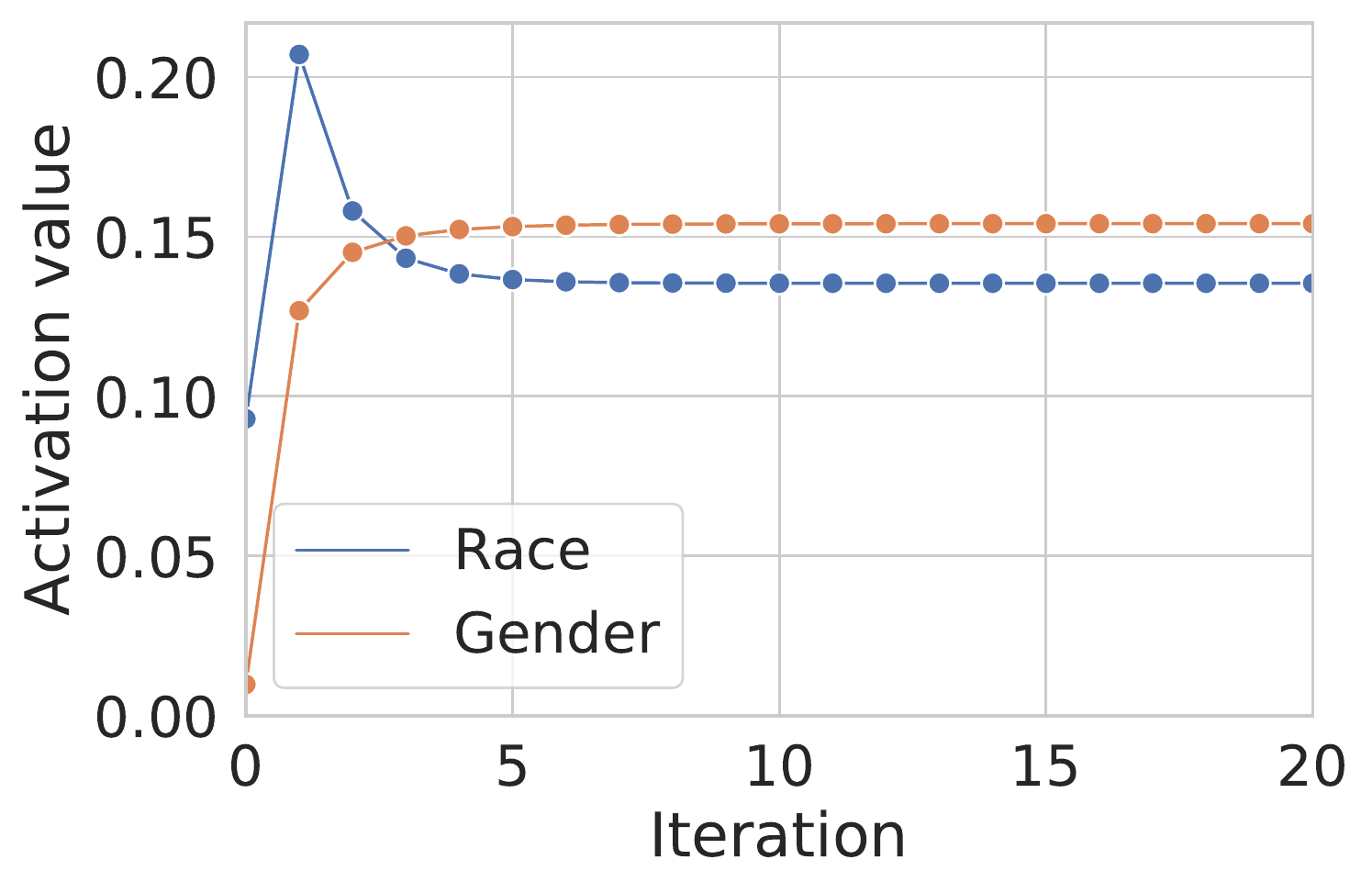}
	\caption{$\phi=0.4$}
	\end{subfigure}
	
	\begin{subfigure}{0.49\textwidth}
	\center
	\includegraphics[width=\textwidth]{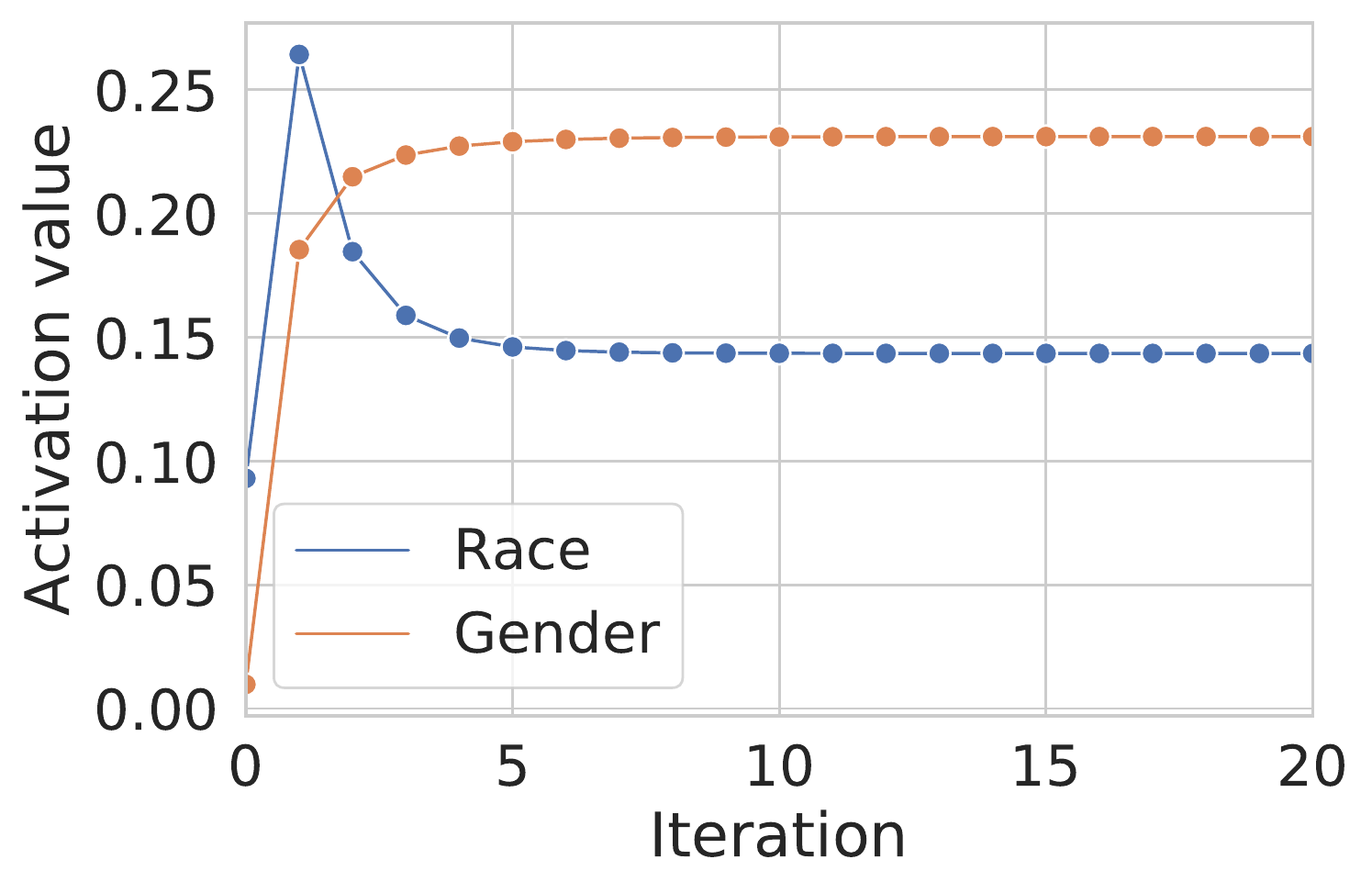}
	\caption{$\phi=0.6$}
	\end{subfigure}
	\begin{subfigure}{0.49\textwidth}
	\center
	\includegraphics[width=\textwidth]{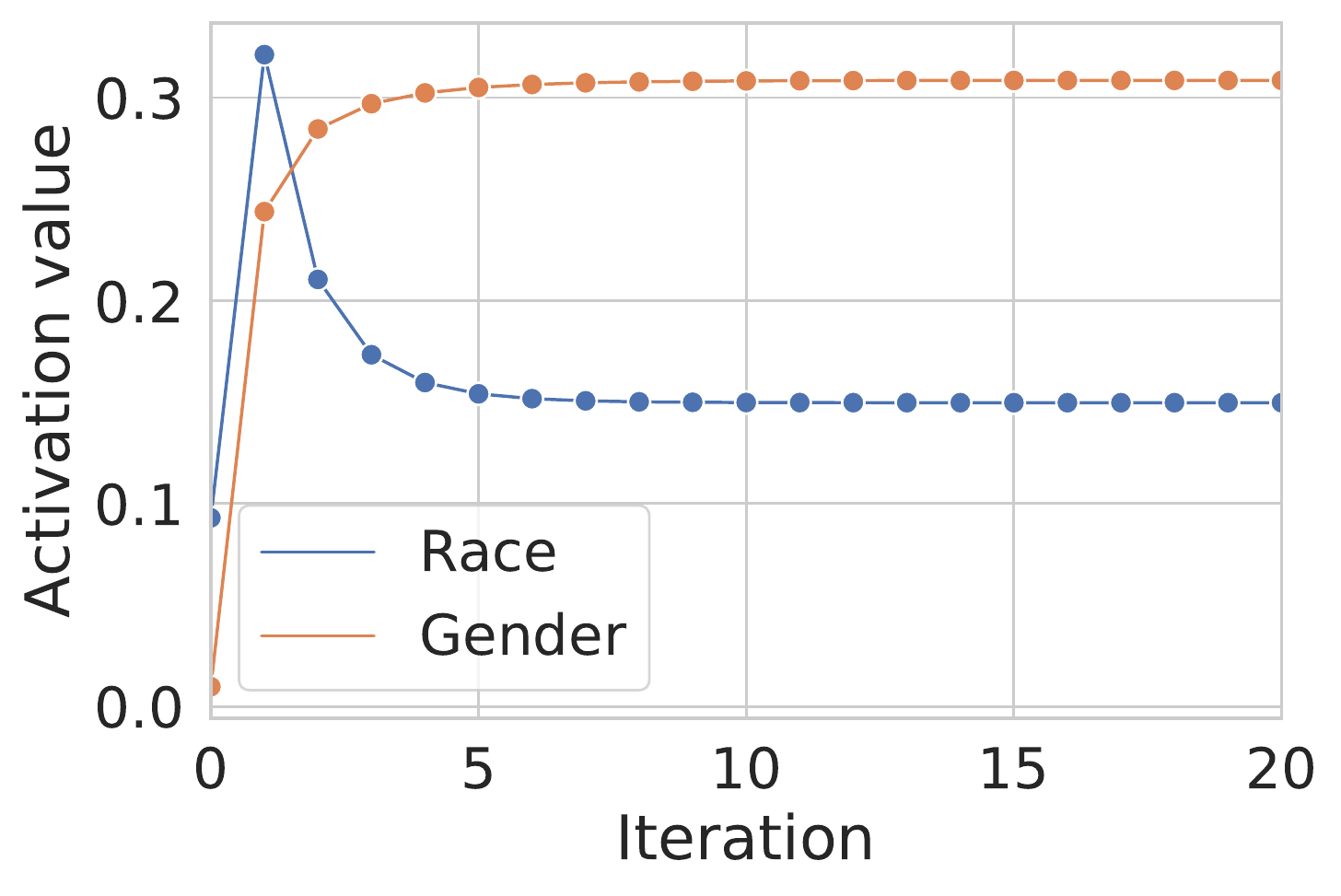}
	\caption{$\phi=0.8$}
	\end{subfigure}
	
	\captionsetup{justification=justified}
	\caption{Activation values of neurons denoting protected features for a negative instance in the Adult dataset for different $\phi$ values..}
\label{fig:fcm_bad_adult}
\end{figure*}

We repeat our analysis using the SHAP values corresponding to the negative instance as an initialization vector in our FCM model. Figure \ref{fig:fcm_bad_adult} plots the activation values of the protected features \textit{race} (F9) and \textit{sex} (F10). We can observe that the initial SHAP value associated with \textit{sex} (F10) is close to zero, while the initial value for \textit{race} (F9) is almost 0.1. Again, after a few iterations, the same pattern as before emerges: \textit{sex} (F10) is twice are important as \textit{race} (F9) thus implicitly biasing the rest of the features in the complex system. 

\begin{figure*}[!htbp]
\center

    \begin{subfigure}{0.8\textwidth}
	\center
	\includegraphics[width=\textwidth]{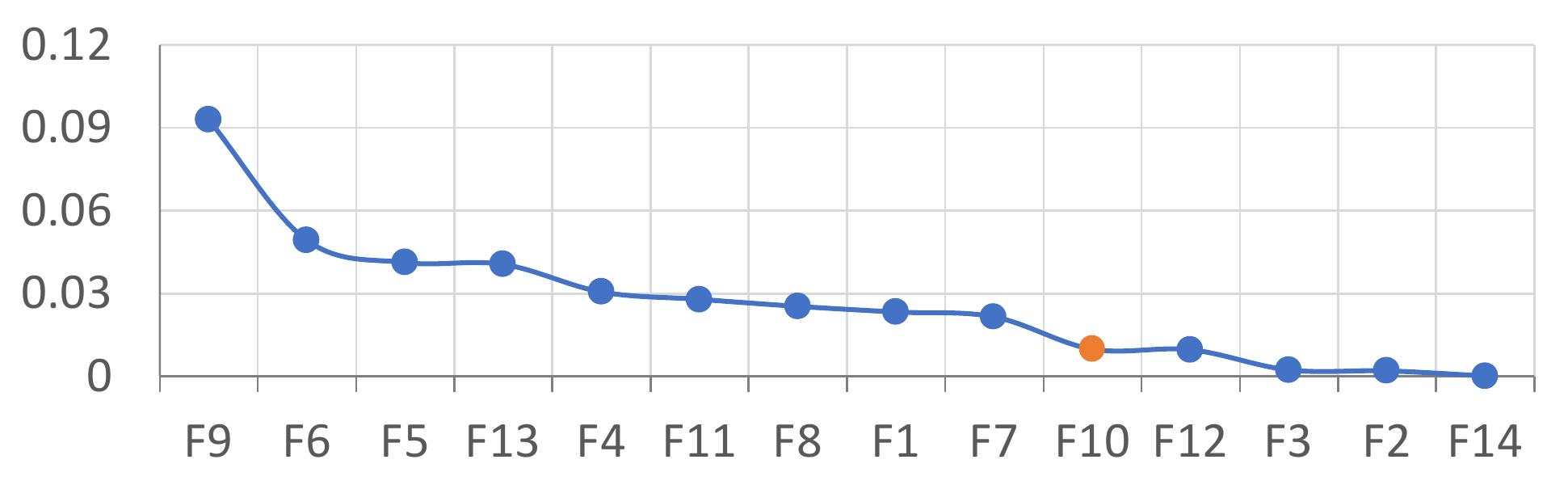}
	\caption{SHAP values}
	\end{subfigure}
	\begin{subfigure}{0.8\textwidth}
	\center
	\includegraphics[width=\textwidth]{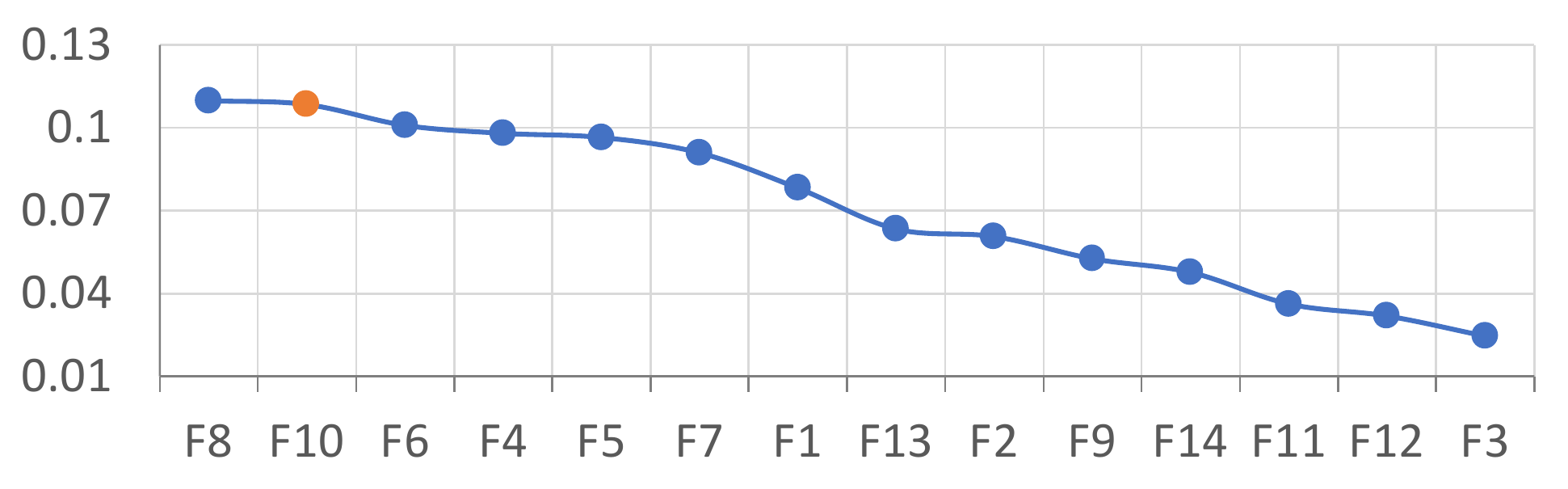}
	\caption{FCM values}
	\end{subfigure}
	
	\captionsetup{justification=justified}
	\caption{Absolute SHAP values and bias scores computed by the FCM model for the randomly selected negative instance of the Adult dataset. In this case, the protected feature \textit{sex} (F10) is deemed not very relevant according to SHAP, but it does involve a significant amount of bias.}
\label{fig:comparison_adult}
\end{figure*}

In an effort to assess whether the protected features rank comparably w.r.t. the SHAP values and the outputs of the FCM model, the reader can refer to Figure \ref{fig:comparison_adult} which displays the aggregated SHAP values and the normalized activation values produced by the FCM model in the last iteration for the negative instance when $\phi = 0.8$. This figure shows that the protected feature \textit{sex} (F10) is one of the least relevant features according to the SHAP values. However, our FCM ranks \textit{sex} (F10) as the second most relevant feature thus implying that it is implicitly influencing all other features to a high extent. At the same time, the exact opposite trend is observed regarding the protected feature \textit{race} (F9) thus proving that a linear method, such as SHAP fails to capture the feedback loops and implicit interactions as encoded in the data.

The simulations provide evidence that, if a protected feature seems to be important by the SHAP method, the rest of the unprotected features do not necessarily have to partially encode bias related to that particular protected feature. This is a situation where a protected feature is actually independent relative to this negative instance (namely \textit{gender} here). On the other hand, \textit{race} confirms our main finding: it is a moderately important feature according to SHAP, but its implicit influence in the system is high.

\section{Conclusions}
\label{sec:conclusions}

This paper developed a methodology to study the relationship between feature importance and implicit bias. Firstly, a classifier was built and optimized in order to predict unseen instances. Secondly, we built a recurrent neural network devoted to quantifying implicit bias from the statistical association patterns between the features. Finally, SHAP feature importance values associated with testing instances were used to trigger the reasoning mechanism.

The simulations using the German Credit and the Adult datasets as case studies showed that there might be situations where protected features are not deemed relevant, yet the amount of implicit bias against them was found significant. In this sense, feature importance and the measures for quantifying explicit bias (such as the measure proposed in \cite{Napoles2022c}) fail to capture the extent to which unprotected features encode the implicit bias patterns. The FCM model presented in \cite{Napoles2022a} tackles this limitation by exploiting the statistical associations between variables. However, the strategy to activate the network relied on expert knowledge, which is often difficult to acquire and quantify. In our approach, such knowledge is replaced with the SHAP values, which provide an elegant alternative to quantify the extent to which each variable is active in the model. In this way, we can measure the extent to which an implicitly biased unprotected feature influences the prediction for a single instance.

Another aspect studied in our paper was the extent to which encoding the numeric features (when analyzing the association between categorical and numerical features) would change the bias patterns. With this aim, we contrasted our results with the simulations reported in \cite{Napoles2022a} and \cite{Napoles2022c} where more bias against gender than age was found. In our experiments, we observed more discrimination against age than gender after detecting the groups automatically using a clustering algorithm. These differences raise concerns about the consistency of existing approaches for detecting bias since a malicious decision-maker could select one approach over another to justify biased decisions.

\bibliographystyle{splncs04}
\bibliography{references}

\begin{thebibliography}{10}
\providecommand{\url}[1]{\texttt{#1}}
\providecommand{\urlprefix}{URL }
\providecommand{\doi}[1]{https://doi.org/#1}

\bibitem{alves2020fixout}
Alves, G., Bhargava, V., Bernier, F., Couceiro, M., Napoli, A.: Fixout: an
  ensemble approach to fairer models (2020),
  \url{https://hal.archives-ouvertes.fr/hal-03033181}

\bibitem{bezdek2013pattern}
Bezdek, J.C.: Pattern recognition with fuzzy objective function algorithms.
  Springer New York, NY (2013). \doi{10.1007/978-1-4757-0450-1}

\bibitem{Bezdek1984}
Bezdek, J.C., Ehrlich, R., Full, W.: Fcm: The fuzzy c-means clustering
  algorithm. Computers \& Geosciences  \textbf{10}(2),  191--203 (1984).
  \doi{10.1016/0098-3004(84)90020-7}

\bibitem{Breiman2001}
Breiman, L.: Random forests. Machine Learning  \textbf{45}(1),  5--32 (2001).
  \doi{10.1023/A:1010933404324}

\bibitem{cesaro2019measuring}
Cesaro, J., Gagliardi~Cozman, F.: Measuring unfairness through game-theoretic
  interpretability. In: Joint European Conference on Machine Learning and
  Knowledge Discovery in Databases. pp. 253--264. Springer (2019).
  \doi{10.1007/978-3-030-43823-422}

\bibitem{cramer2016mathematical}
Cram{\'e}r, H.: Mathematical Methods of Statistics. Princeton University Press
  (2016). \doi{10.1515/9781400883868}

\bibitem{Dua2019}
Dua, D., Graff, C.: {UCI} machine learning repository (2017),
  \url{https://archive.ics.uci.edu/ml/datasets/statlog+(german+credit+data)}

\bibitem{fang2020achieving}
Fang, B., Jiang, M., Cheng, P.y., Shen, J., Fang, Y.: Achieving outcome
  fairness in machine learning models for social decision problems. In:
  Bessiere, C. (ed.) Proceedings of the Twenty-Ninth International Joint
  Conference on Artificial Intelligence, {IJCAI-20}. pp. 444--450.
  International Joint Conferences on Artificial Intelligence Organization
  (2020). \doi{10.24963/ijcai.2020/62}

\bibitem{hajian2012methodology}
Hajian, S., Domingo-Ferrer, J.: A methodology for direct and indirect
  discrimination prevention in data mining. IEEE transactions on knowledge and
  data engineering  \textbf{25}(7),  1445--1459 (2012).
  \doi{10.1109/TKDE.2012.72}

\bibitem{hickey2020fairness}
Hickey, J.M., Di~Stefano, P.G., Vasileiou, V.: Fairness by explicability and
  adversarial shap learning. In: Machine Learning and Knowledge Discovery in
  Databases: European Conference, ECML PKDD 2020, Ghent, Belgium, September
  14–18, 2020, Proceedings, Part III. p. 174–190. Springer-Verlag (2020).
  \doi{10.1007/978-3-030-67664-311}

\bibitem{Kohavi1996ScalingUT}
Kohavi, R.: Scaling up the accuracy of naive-bayes classifiers: A decision-tree
  hybrid. In: Proceedings of the Second International Conference on Knowledge
  Discovery and Data Mining. p. 202–207. AAAI Press (1996)

\bibitem{Kosko1986}
Kosko, B.: Fuzzy cognitive maps. International Journal of Man-Machine Studies
  \textbf{24}(1),  65--75 (1986). \doi{10.1016/s0020-7373(86)80040-2}

\bibitem{lundberg2020local2global}
Lundberg, S.M., Erion, G., Chen, H., DeGrave, A., Prutkin, J.M., Nair, B.,
  Katz, R., Himmelfarb, J., Bansal, N., Lee, S.I.: From local explanations to
  global understanding with explainable {AI} for trees. Nature Machine
  Intelligence  \textbf{2}(1),  2522--5839 (2020).
  \doi{10.1038/s42256-019-0138-9}

\bibitem{NIPS2017_7062}
Lundberg, S.M., Lee, S.I.: A unified approach to interpreting model
  predictions. In: Guyon, I., Luxburg, U.V., Bengio, S., Wallach, H., Fergus,
  R., Vishwanathan, S., Garnett, R. (eds.) Advances in Neural Information
  Processing Systems 30, pp. 4765--4774. Curran Associates, Inc. (2017).
  \doi{10.5555/3295222.3295230}

\bibitem{mehrabi2021survey}
Mehrabi, N., Morstatter, F., Saxena, N., Lerman, K., Galstyan, A.: A survey on
  bias and fairness in machine learning. ACM Computing Surveys (CSUR)
  \textbf{54}(6),  1--35 (2021). \doi{10.1145/3457607}

\bibitem{meng2022interpretability}
Meng, C., Trinh, L., Xu, N., Enouen, J., Liu, Y.: Interpretability and fairness
  evaluation of deep learning models on mimic-iv dataset. Scientific Reports
  \textbf{12}(1),  1--28 (2022). \doi{10.1038/s41598-022-11012-2}

\bibitem{nagelkerke1991note}
Nagelkerke, N.J., et~al.: A note on a general definition of the coefficient of
  determination. Biometrika  \textbf{78}(3),  691--692 (1991).
  \doi{10.1093/biomet/78.3.691}

\bibitem{Napoles2022a}
N\'apoles, G., Grau, I., Concepci\'on, L., {Koutsoviti Koumeri}, L., Papa,
  J.P.: Modeling implicit bias with fuzzy cognitive maps. Neurocomputing
  \textbf{481},  33--45 (2022). \doi{10.1016/j.neucom.2022.01.070}

\bibitem{Napoles2022c}
N\'apoles, G., {Koutsoviti Koumeri}, L.: A fuzzy-rough uncertainty measure to
  discover bias encoded explicitly or implicitly in features of structured
  pattern classification datasets. Pattern Recognition Letters  \textbf{154},
  29--36 (2022). \doi{doi.org/10.1016/j.patrec.2022.01.005}

\bibitem{Napoles2022b}
N\'apoles, G., Salgueiro, Y., Grau, I., Espinosa, M.L.: Recurrence-aware
  long-term cognitive network for explainable pattern classification. IEEE
  Transactions on Cybernetics pp. 1--12 (2022). \doi{10.1109/TCYB.2022.3165104}

\bibitem{napoles2020}
N{\'a}poles, G., Salmeron, J.L., Froelich, W., Falcon, R., Leon~Espinosa, M.,
  Vanhoenshoven, F., Bello, R., Vanhoof, K.: Fuzzy cognitive modeling:
  Theoretical and practical considerations. In: Intelligent Decision
  Technologies 2019, pp. 77--87. Springer (2020)

\bibitem{ntoutsi2020bias}
Ntoutsi, E., Fafalios, P., Gadiraju, U., Iosifidis, V., Nejdl, W., Vidal, M.E.,
  Ruggieri, S., Turini, F., Papadopoulos, S., Krasanakis, E., et~al.: Bias in
  data-driven artificial intelligence systems—an introductory survey. Wiley
  Interdisciplinary Reviews: Data Mining and Knowledge Discovery
  \textbf{10}(3),  e1356 (2020). \doi{10.1002/widm.1356}

\bibitem{rovine199714th}
Rovine, M.J., Von~Eye, A.: A 14th way to look at a correlation coefficient:
  Correlation as the proportion of matches. The American Statistician
  \textbf{51}(1),  42--46 (1997). \doi{10.1080/00031305.1997.10473586}

\bibitem{shapley1953value}
Shapley, L.S.: A value for n-person games. Contributions to the Theory of Games
   \textbf{2}(28),  307--317 (1953). \doi{10.1515/9781400881970-018}

\bibitem{zhang2018fairness}
Zhang, J., Bareinboim, E.: Fairness in decision-making—the causal explanation
  formula. In: Proceedings of the AAAI Conference on Artificial Intelligence.
  vol.~32 (2018). \doi{10.1609/aaai.v32i1.11564}

\bibitem{vzliobaite2017measuring}
{\v{Z}}liobait{\.e}, I.: Measuring discrimination in algorithmic decision
  making. Data Mining and Knowledge Discovery  \textbf{31}(4),  1060--1089
  (2017). \doi{10.1007/s10618-017-0506-1}

\end{thebibliography}

\end{document}